\documentclass[10pt,twocolumn,letterpaper]{article}

\usepackage[pagenumbers]{cvpr} 

\usepackage{times}
\usepackage{epsfig}
\usepackage{graphicx}
\usepackage{amsmath}
\usepackage{amssymb}
\usepackage{indentfirst}
\usepackage{booktabs}

\usepackage{amssymb}
\usepackage{mathtools}
\usepackage{subcaption}
\usepackage{multirow}
\usepackage{adjustbox}
\usepackage{xcolor}
\usepackage{enumitem}
\usepackage{booktabs}
\usepackage{pifont}
\usepackage{dsfont}
\usepackage{xstring}
\usepackage{textcomp}
\usepackage{algorithm, algorithmic}
\usepackage[normalem]{ulem}
\DeclareMathOperator*{\argmax}{arg\,max}
\DeclareMathOperator*{\argmin}{arg\,min}

\usepackage[accsupp]{axessibility}

\usepackage{caption}
\newcommand{\algname}{{HyperFD}}
\newcommand{\diff}[1]{\IfSubStr{#1}{+}{{\color{blue}#1}}{{\color{red}#1}}}
\newcommand{\meanstd}[2]{#1{\small $\pm$}#2}

\newcommand{\highlight}[1]{\noindent\textbf{#1}}

\usepackage[pagebackref,breaklinks,colorlinks]{hyperref}

\usepackage[capitalize]{cleveref}
\crefname{section}{Sec.}{Secs.}
\Crefname{section}{Section}{Sections}
\Crefname{table}{Table}{Tables}
\crefname{table}{Tab.}{Tabs.}

\def\cvprPaperID{9981} 
\def\confName{CVPR}
\def\confYear{2022}

\begin{document}

\title{Privacy-preserving Online AutoML for Domain-Specific Face Detection}

\author{Chenqian Yan$^{1 \dag}$\thanks{Work done as an intern at MSRA. $^\dag$ Equal contribution.}  \quad Yuge Zhang$^{1 \dag}$ \quad Quanlu Zhang$^1$ \quad Yaming Yang$^1$ \\ Xinyang Jiang$^1$ \quad Yuqing Yang$^1$ \quad Baoyuan Wang$^2$ \\
Microsoft Research$^1$ \quad Xiaobing.ai$^2$  \\
{\tt\footnotesize im.cqyan@gmail.com,\{yugzhan,quzha,yayaming,xinyangjiang,yuqyang\}@microsoft.com,wangbaoyuan@xiaobing.ai} \\ 
}

\maketitle

\begin{abstract}
Despite the impressive progress of general face detection, the tuning of hyper-parameters and architectures is still critical for the performance of a domain-specific face detector. Though existing AutoML works can speedup such process, they either require tuning from scratch for a new scenario or do not consider data privacy.
To scale up, we derive a new AutoML setting from a platform perspective.
In such setting, new datasets sequentially arrive at the platform, where an architecture and hyper-parameter configuration is recommended to train the optimal face detector for each dataset. This, however, brings two major challenges: (1) how to predict the best configuration for any given dataset without touching their raw images due to the privacy concern? and (2) how to continuously improve the AutoML algorithm from previous tasks and offer a better warm-up for future ones?  We introduce ``\algname{}'', a new privacy-preserving online AutoML framework for face detection. At its core part, a novel meta-feature representation of a dataset as well as its learning paradigm is proposed. Thanks to \algname{}, each local task (client) is able to effectively leverage the learning ``experience'' of previous tasks without uploading raw images to the platform; meanwhile, the meta-feature extractor is continuously learned to better trade off the bias and variance. Extensive experiments demonstrate the effectiveness and efficiency of our design.

\end{abstract}

\vspace{-1em}

\section{Introduction}
\label{sec:intro}



Face detection \cite{deng2019retinaface,liao2015fast,zhang2018faceboxes,zhang2016joint,vesdapunt2021crface} is one of the most fundamental problems in computer vision. Although, rapid progress has been made lately for the general cases, bespoken face detection models are still in high-demand for domain-specific scenarios. This is because, the challenges for detecting faces from an outdoor surveillance camera might be different from a panoramic indoor fish-eye camera ~\cite{fu2019fddb360}; likewise, the challenges for detecting occluded faces (\eg, masks~\cite{jiang2020retinamask}) are also quite different from selfie faces captured by cellphone cameras ~\cite{mahbub2017partial}. Therefore, extensive manual parameter tuning and large computing resource is required in order to obtain the best specialized model for each domain. 
To scale up the scenarios, we see a clear industry demand of building shared AI model training platform so as to leverage pretrained representation from other relevant tasks. For example, Microsoft Custom Vision\cite{MCV} can train specialized object detection models given a set of user-uploaded images for engineers without deep learning background. This, however, comes at the cost of sacrificing the face data privacy, for face detection models. Similarly, other AutoML tools (\eg, NNI~\cite{NNI}) either do not consider the data privacy or still require tuning from scratch for a new scenario, which is not secure and scalable.

\begin{figure}[tbp]
    \centering
    \includegraphics[width=\linewidth]{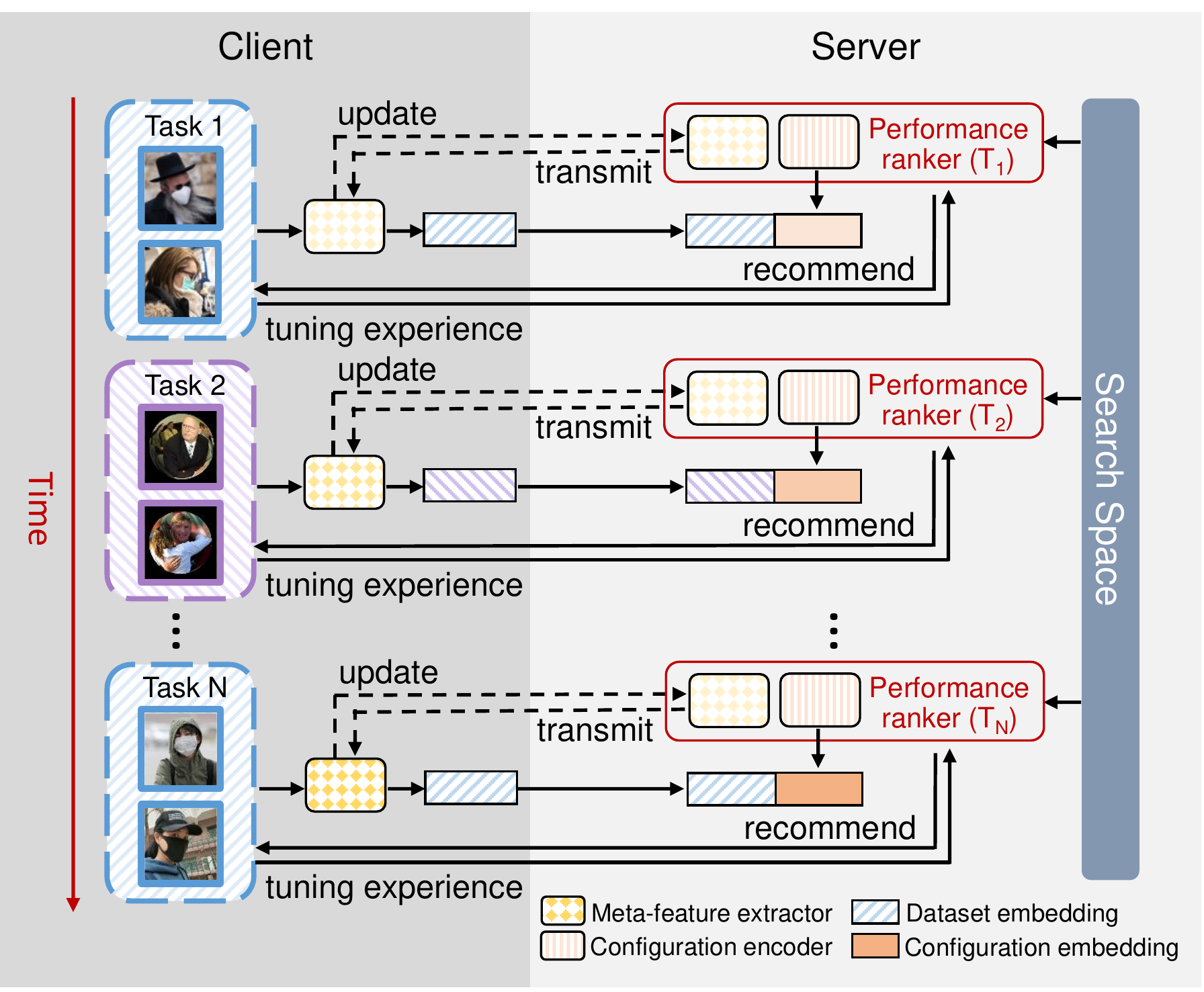}
    \caption{Overview of \algname{} framework, which aims to build a shared AutoML platform that enables exchanging tuning experience among customers, without access to customers' raw datasets. The performance ranker consists of meta-feature extractor and configuration encoder. Both are continuously updated to incorporate tuning experience on the latest task.}
    \label{fig:framework}
    \vspace{-2em}
\end{figure}

The training of domain-specific face detection for real-world scenarios requires a new problem setup from a platform perspective, where the platform receives new datasets sequentially, and recommends architecture and hyper-parameter configuration to train the optimal face detector for each dataset (corresponds one particular domain). This problem setting brings two major challenges. The first is to effectively predict the best configuration for any incoming dataset under the constraints that the privacy of their raw images are protected. The second is to leverage the ``experience'' from previous tasks\footnote{In this paper, we use ``task'' and ``dataset'' interchangeably.} to continuously improve the AutoML algorithm, such that the platform can better serve future ones. 


To tackle the challenge of privacy protection, we derive a new online AutoML paradigm for face detection, which is called ``\algname{}". 
Specifically, instead of uploading the raw images to the server, each local client only sends the dataset level representations (called ``meta-features'' in the following) to the platform and asks for the best configuration including a network architecture and hyper-parameters. The meta-feature is designed to encode the overall statistics and general attributes of a given dataset. The platform maintains a learnable performance ranker that selects top-$k$ optimal training configurations from the hyper-parameter/architecture search space based on the dataset meta-feature. Finally, after obtaining the configurations from the platform, the face detector is then trained locally on each client. In this way, the platform only sees dataset meta-features and the testing performance, which effectively protect the training data privacy. 
Figure \ref{fig:framework} gives an overview of our \algname{} framework. Note that, although it is federated, HyperFD conducts the actual training task locally and there is no global aggregation is needed, which is different from traditional federated learning \cite{McMahan2017CommunicationEfficientLO}.

To tackle the second challenge and make HyperFD more generalizable for unseen scenarios, we ask the meta-feature to be updated continuously with the new dataset, yet properly borrow the ``experience'' from previously trained tasks. Due to the fact that there is no way to access the raw data, we integrate a novel meta-feature transformation module that builds a mapping between the current meta-feature space and previous feature space. Intuitively, this mapping will help make similar distributed historical tasks play more influence in ranking the final configuration for the new task. To summarize, we make the following contributions:

\vspace{-0.5em}

\begin{itemize}
    \item We introduce privacy-preserving online AutoML for face detection, which is a new problem setting from platform perspective.
    \item We propose a novel meta-feature extractor to build better dataset level representations, which is trained continuously without touching the raw face images.
    \item Extensive experiments show the superior performance of our approach. We will also release the benchmark and source code to facilitate future research.
\end{itemize}

\section{Related works}

\highlight{Transferable AutoML.} 
Early works of transferable AutoML addressed the problem from a multi-task collaborative tuning perspective~\cite{bardenet2013collaborative, swersky2013multi, fusi2018probabilistic}, in the hope that running multiple AutoML tasks simultaneously will help each other achieve better results.
The underlying techniques include surrogate-based ranking~\cite{bardenet2013collaborative}, multi-task Gaussian processes~\cite{swersky2013multi} and probabilistic matrix factorization~\cite{fusi2018probabilistic}. 
A later stream of researches focus on the ``warm-start'' setting, which is to recommend a good starting point for a new AutoML task based on previous experiments~\cite{feurer2015initializing,yang2019oboe,kim2017learning,mittal2020hyperstar}. 
However, from a platform perspective, a more realistic setting would be, training tasks arrive sequentially and one needs to search the best configuration for a coming dataset. Xue \etal~\cite{xue2019transferable} firstly refers to this setting as ``AutoML under \emph{online setting}''. 
Although a few attempts have been made to formulate and solve such problem~\cite{xue2019transferable,xu2021nasoa}, they still neglected the essential constraints of privacy and assumed all datasets to be directly accessible to the algorithm, which is infeasible especially for sensitive datasets like face.

\highlight{Dataset meta-feature.}
Dataset meta-feature (or representation) is shown to be crucial to the performance of transferable AutoML~\cite{jomaa2021dataset2vec}.
The most straight-forward and earliest dataset representation is based on descriptive statistics of a dataset~\cite{feurer2020autosklearn,michie1994machine, kalousis2002algorithm,yogatama2014efficient,bardenet2013collaborative}, \eg, number of images. 
More advanced approaches include the usage of pretrained models' performances on unseen datasets~\cite{feurer2015initializing,xue2019transferable,xiao2021amortized} along with their landscape~\cite{achille2019task2vec}.  
However, these methods are heuristically designed and not directly aware of the end-to-end goal of AutoML. An alternative way is to optimize deep features of neural network~\cite{kim2017learning,wong2018transfer,mittal2020hyperstar,jomaa2021dataset2vec} in an end-to-end manner. Although effective, this has not become the mainstream and recent works targeting at online setting~\cite{xue2019transferable,xu2021nasoa} are still using heuristically designed representations. 
Moreover, most works are limited on image classification and tabular dataset for their evaluation. To the best of our knowledge, we are the first to extract meta-features for face detection datasets.

\highlight{Continual learning.}
Continual learning is a scenario where a single neural network needs to sequentially learn a series of tasks. The crucial challenge is catastrophic forgetting that parameters or semantics 
learned for the past tasks drift to the direction of new tasks. To alleviate such issue, regularization on gradients~\cite{kirkpatrick2017overcoming,zenke2017continual}, designs of dynamic architectures~\cite{yoon2017lifelong,xu2018reinforced,yoon2019scalable}, replay of previous training data~\cite{li2017learning,shin2017continual,rebuffi2017icarl} are usually needed.
Despite the various techniques, joint training on all history data is still considered the upper bound for continual learning~\cite{van2019three}.
Recent researches proposed federated continual learning~\cite{yoon2021federated,jiang2021fedspeech}, putting privacy into consideration, but they are essentially different from us. Their clients aggregate parameters, while we propose to share meta-level knowledge of AutoML experience. Moreover, their solution is designed for Task-IL scenario~\cite{van2019three} and not applicable to ours.

\newcommand{\Expect}{\mathbb{E}}
\newcommand{\SSpace}{\mathcal{C}}
\newcommand{\SPerf}{\mathcal{S}}
\newcommand{\SData}{\mathcal{D}}
\newcommand{\Doff}{\mathcal{D}_{\textrm{offline}}}
\newcommand{\SHyper}{\mathcal{H}}
\newcommand{\FF}{\mathcal{F}}
\newcommand{\SExt}{\mathcal{E}}
\newcommand{\hyper}{\mathrm{hyper}}
\newcommand{\aug}{\mathrm{aug}}
\newcommand{\past}{\mathrm{past}}
\newcommand{\Loss}{\mathcal{L}}
\newcommand{\AP}{\mathrm{AP}}
\newcommand{\Gi}{\ensuremath{G^{(i)}}}
\newcommand{\Gj}{\ensuremath{G^{(j)}}}
\newcommand{\Gt}{\ensuremath{G^{(t)}}}
\newcommand{\Zb}{\textbf{Z}}
\newcommand{\Zbi}{\ensuremath{\mathbf{Z}^{(i)}}}
\newcommand{\Zbt}{\ensuremath{\mathbf{Z}^{(t)}}}
\newcommand{\cui}{\ensuremath{c_{u_i}}}
\newcommand{\dvi}{\ensuremath{d_{v_i}}}
\newcommand{\phivi}{\ensuremath{\phi_{v_i}}}
\DeclarePairedDelimiter{\floor}{\lfloor}{\rfloor}

\section{\algname{}}



\subsection{Framework overview}
\autoref{fig:framework} illustrates an overview of our \algname{} framework, where the domain-specific training tasks for face detection sequentially arrive. For a new task, meta-feature extractor maintained on the server is transmitted to the client. The client extracts features from the face dataset with the extractor, and sends the features back to the server without disclosing the raw images. The server maintains a search space which has various model architectures and hyper-parameters for face detection. A performance ranker predicts the rank of the configurations on that client's task based on its meta-feature and suggests several configurations to the client. The client verifies the performance of those configurations on its dataset, and informs the performance to the server. 
Then the server updates the performance ranker with the newly collected data. This is a continual learning process with tuning experience on new face detection tasks continuously arriving. The problem is that raw datasets are not available on the server and every client becomes unreachable after its task is done. Thus, we design a meta-feature transformation module. It continuously projects the features extracted with old-version extractors to the feature space of the latest extractor, so that the updates could leverage all the historical data to prevent forgetting. 
The loss function combines ranking loss, regularization, and synaptic intelligence to guarantee steady improvement of the whole framework.

\subsection{Performance ranker}
\label{sec:performance-ranker}

Performance ranker is the basic building block of \algname{} framework. It ranks configurations from a search space for each dataset. The search space $\SSpace$ contains $| \SSpace | = M$ different configurations, the $k$-th of which is denoted as $c_k$. We use $\SData = \{d_1, d_2, \ldots, d_N\}$ to denote the datasets of  face detection tasks, where $d_t$ corresponds to the $t$-th task. Our goal is to learn a performance ranker that, for any pair of $c_k$ and $d_t$, it predicts a score (\eg, AP@50).


We formulate the ranker as a differentiable function $F(c_k, d_t; \theta_F)$ parameterized by $\theta_F$. $F$ consists of two key components: a configuration encoder $H(c_k; \theta_H)$ that maps any configuration in $\SSpace$ into a fixed-length vector, and a dataset meta-feature extractor $G(d_t; \theta_G)$ that extracts semantic information from a dataset to generate a fixed-length vector. 
$G$ and $H$ are learned such that the configuration embedding and dataset embedding lie in the same embedding space, and we measure their similarity with an inner product. A higher similarity means a better match of an configuration and a dataset, leading to a potentially better performance.
\begin{equation}
F(c_k, d_t; \theta_F) = G(d_t; \theta_G)^\intercal  H(c_k; \theta_H)
\end{equation}

In our framework, the performance ranker is optimized in a supervised learning manner on \emph{tuning experience}.
Here, tuning experience is a meta-dataset consisting of a number of triplets $\{(\cui, \dvi, a_i)\}_{i=1}^S$ ($1 \le u_i \le M$, $1 \le v_i \le N$), where $u_i$ is the index of the configuration evaluated on the $v_i$-th dataset in the $i$-th triplet, $a_i$ is its performance, \ie, evaluation score of a detector trained with $\cui$ on $\dvi$. 

\subsubsection{Meta-feature extractor}

 \begin{figure}[tbp]
\centering
\includegraphics[width=\linewidth]{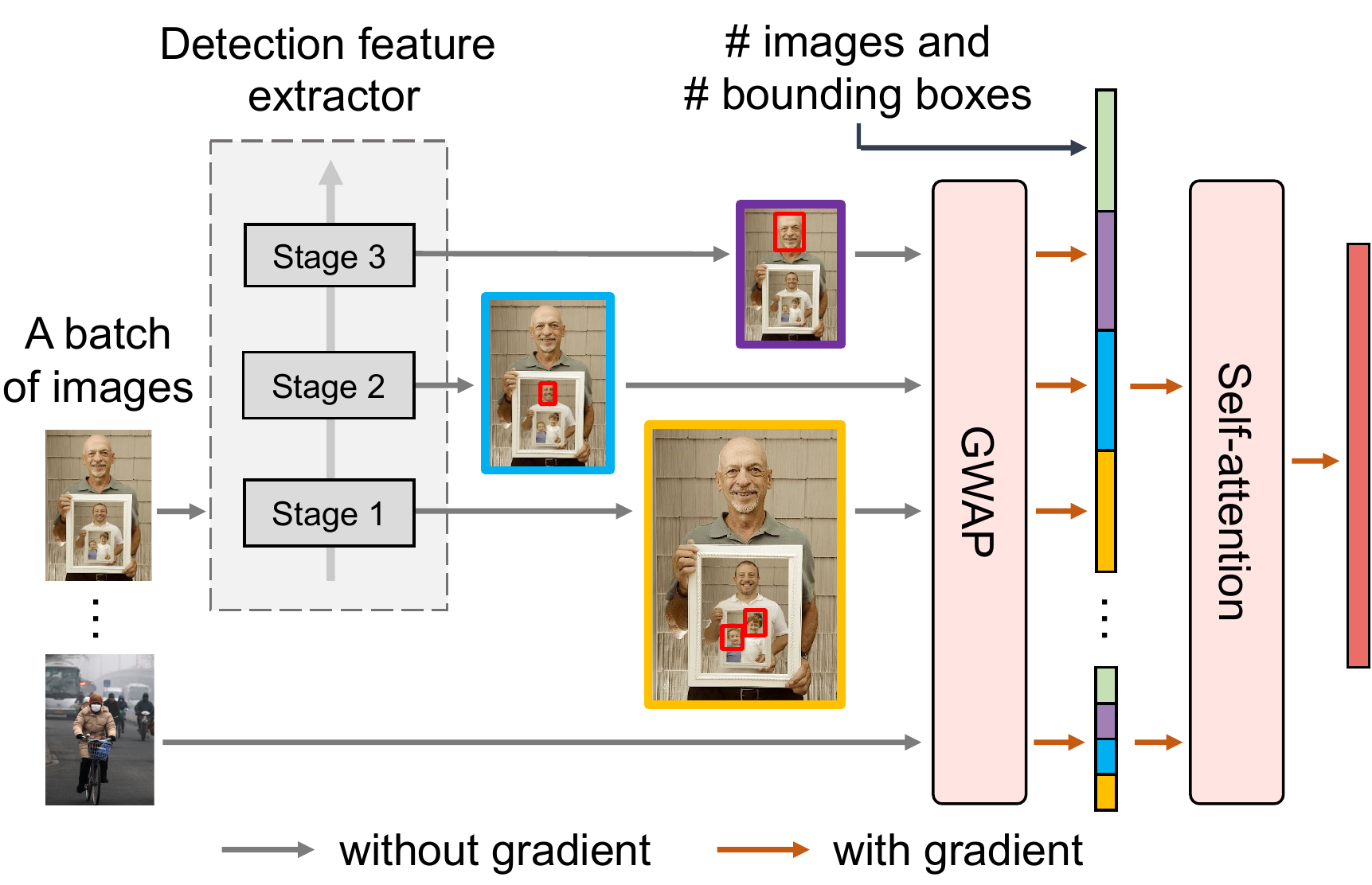}
\caption{Architecture of meta-feature extractor.}
\vspace{-0.5em}
\label{fig:framework-meta-extractor}
\vspace{-0.5em}
\end{figure}


The key challenge to design a performance ranker is how to effectively extract meta-features from a dataset. Inspired by previous works on image classification~\cite{mittal2020hyperstar,kim2017learning}, we follow the intuition that semantic information from the raw images (\eg, statistical information of images and annotations, vision features extracted with a pretrained model) could decently convey the characteristics of a dataset. However, face detection is more complex, which spans multi-scale anchors combined with per-anchor classification and regression. Thus, we design a novel hierarchical feature extraction approach starting from anchor-level, followed by a series of aggregations to generate a high-level dataset feature. The overall architecture is shown in \autoref{fig:framework-meta-extractor}.


\highlight{Anchor-level.} To handle raw images, we first feed them into a pretrained face detector, RetinaFace~\cite{deng2019retinaface}. We use the feature pyramid generated by context modules (\ie SSH~\cite{najibi2017ssh}), which consists of three feature maps downsampled by 8, 16 and 32 respectively. These feature maps are responsible for detecting small, medium, and large faces in the original detector design. We call those feature maps anchor-level features because every pixel on the feature map corresponds to one or several anchors in detection tasks. Apart from features extracted by detector, we also attach the matching ground-truth bounding boxes to each anchor to enrich the information. The feature map for the $k$-th stage is denoted with $\textrm{det}^k(I)$, whose height is $H_k$ and width is $W_k$, where $I$ is a preprocessed image,

\highlight{Image-level.} To aggregate anchor-level features into image-level, each of the three feature maps is fed into a Global Weighted Average Pooling (GWAP)~\cite{qiu2018global}, which in our case is to balance the weights for unbalanced positive and negative samples in face detection. We reweight the anchors such that different levels of IoU matching ratios contribute equally to the results. Concretely, for $k=1,2,3$, we group the anchors on $\textrm{det}^k(I)$ according to levels of IoU matching ratios into positive anchors, negative anchors, and ignored anchors~\cite{chen2019mmdetection}, and average the anchors within groups before a total average:
\begin{equation}
\overline{\textrm{det}}^k(I) = \frac{1}{3} \left(\overline{\textrm{det}}^k_{\textrm{pos}}(I) + \overline{\textrm{det}}^k_{\textrm{neg}}(I) + \overline{\textrm{det}}^k_{\textrm{ign}}(I)  \right)
\end{equation}
where $\overline{\textrm{det}}^k_{\textrm{pos}}$, $\overline{\textrm{det}}^k_{\textrm{neg}}$, $\overline{\textrm{det}}^k_{\textrm{ign}}$ are the average of the stage-$k$ feature map over positive, negative and ignored anchors respectively.
The feature for one image is then a concatenation of $\overline{\textrm{det}}^k(I)$ for $k=1, 2, 3$. 
Another vector of high-level descriptive statistics of the dataset~\cite{yogatama2014efficient} is concatenated, which, in our case, are number of images and bounding boxes of the whole dataset. This is needed because such information is missing in the context of a batch of images. The information is injected at the image level so that the follow-up modules can fuse it with other visual semantics. 

\highlight{Dataset-level.} This level aggregates the features of all the images of a dataset to capture the distribution of the images' features. We propose to use self-attention (\ie, a transformer encoder layer~\cite{vaswani2017attention}) to extract the distribution semantic.
The rationale behind this is that the optimal configuration suited for one task is mostly correlated to the distribution of the dataset. 
Specifically, positional embedding is not used because the sequence of images is permutation invariant. After the features fused among images, an average pooling is used to obtain a dataset-level feature, \ie, meta-feature of the dataset.


\subsubsection{Configuration encoder}

Different types of search spaces prefer different configuration encoders. For a hyper-parameter search space which consists of one or more categorical variables, Multi-layer Perceptron is a simple yet effective choice. For neural architecture search space, a more sophisticated encoder is desired, so as to handle the complexity in neural networks. There is recently a growing popularity to use Graph Neural Networks as the performance predictor in NAS~\cite{wen2020neural,dudziak2020brp,siems2020bench}, because neural networks are essentially graphs. To be specific, we use GIN~\cite{xu2018powerful} as the configuration encoder for neural architectures due to its superiority over other GNNs\cite{siems2020bench}.

\subsection{Privacy-preserving continual learning}
\label{sec:privacy-continual}

In \algname{}, performance ranker runs in a scenario that raw datasets on the client side are not accessible by the server and the client is only reachable during its own task. However, continuously improving performance ranker requires joint training on experience of both incoming new tasks and old ones. This brings two new challenges to continual learning. Firstly, the server only stores meta-features rather than raw datasets. Secondly, the meta-features are generated with different versions of meta-feature extractor due to continual learning.  
To address those challenges, we design a novel meta-feature transformation module combined with three loss functions, which guarantees stable and effective update of the performance ranker.

\subsubsection{Meta-feature transformation module}

Our performance ranker can be naturally decomposed into $G$ and $H$. $G$ is executed and updated on the client side, thus, preserves privacy of the client. The updated $G$ is sent back to the server. As $G$ keeps evolving along with incoming tasks, we introduce $t$ to denote the $t$-th task (assuming tasks arrive sequentially). The $t$-th task has dataset $d_t$, meta-feature $\Gt (d_t; \theta_G^{(t)})$, where $\theta_G^{(t)}$ is the weight of the meta-feature extractor in the state of finishing $t$-th task. We use $\Gt (d_t)$ for short in the rest of this paper. The tuning experience triplets on the server is $\{\cui, \phivi, a_i\}_{i=1}^S$, where $\phivi$ the meta-feature of $\dvi$, defined as follows:
\begin{equation}
\label{eq:extractor-extended}
\phi_t =
\begin{cases}
G(d_{\tau}; \theta_G), & \text{if } t = \tau \\
G^{(t)}(d_t; \theta_G^{(t)}), & \text{otherwise}
\end{cases}
\end{equation}
where $d_{\tau}$ is the dataset of the current (\ie $\tau$-th) task.



From \autoref{eq:extractor-extended} we can see that, the meta-features of current dataset and the meta-features in the past are extracted with different meta-feature extractors, and lie in different feature spaces. The oracle solution would be that we ask the users to extract the meta-feature again with the current extractor, but it is infeasible in our scenario. To align meta-features to the latest feature space,
we project the meta-features extracted with old meta-feature extractors to the latest feature space using linear mapping. We call this \emph{meta-feature transformation}.

Assume our system is currently ready to serve dataset $d_{\tau}$.
The latest meta-feature extractor is denoted as $G$. We aim to predict $G(d_t)$,  with $\Gt(d_t)$ as input, multiplied by a transformation matrix $\Zbt$ (one matrix for each dataset). It becomes a supervised learning problem to minimize the distance between $\Zbt \Gt(d_t)$ and $G(d_t)$. To learn $\Zbt$, we need a plenty number of pairs  $(\Gt(d), G(d))$, where $d$ is any dataset. It is impossible to collect such data from users, because we cannot expect ``plenty'' of users to be online when serving a new user. Thus, we take $d$ from $\Doff$, a series of datasets prepared in \algname{} framework offline, whose raw data are always accessible. Then $\Zbt$ can be trained via:
\begin{equation}
\label{eq:loss-calib}
\Loss_{\textrm{trans}}(\Zbt) = \sum_{d \in \Doff} \| \Zbt \Gt(d) -  G(d) \|^2
\end{equation}

With the transformation modules $\Zb^{(1)}, \Zb^{(2)}, \ldots, \Zb^{(\tau-1)}$ sufficiently trained, we have $\Zbt \Gt (d) \approx G(d)$ hold for any dataset $d$. We can then extend the definition of performance ranker $F$ to $\FF$, so that it can work on any pair of configurations and meta-features.
\begin{equation}
\label{eq:ranker-extended}
\FF(c_k, \phi_t; \theta_F) =
\begin{cases}
\phi_t^\intercal  H(c_k), & \text{if } t = \tau \\
\left( \mathbf{Z}^{(t)} \phi_t \right)^\intercal H(c_k), & \text{otherwise}
\end{cases}
\end{equation}

In this equation, when updating $\theta_F$ with back propagation, $\theta_G$ can be updated only for the branch of current dataset $d_{\tau}$. On the other hand, for $t< \tau$  there is no gradients, because $\phi_t$ is pre-computed and $\Zbt$ is already optimized. Hence, the joint training on previous experience is essentially adapting the configuration encoder to the feature space of the latest meta-feature extractor.

\subsubsection{Loss functions} 


Next, we introduce the loss function to optimize the performance ranker.

\highlight{Ranking loss.} Our primary loss function is a ranking loss proposed by \cite{zhang2021acenas} that penalizes imperfect ranking of configurations. The loss is calculated on $\{(\cui, \phivi, a_i)\}_{i=1}^{K}$, and is in the form of,
\begin{equation}
\Loss_{\textrm{rank}}(F) = \Expect_{\substack{v_i = v_j \\ a_i > a_j}} [- \Delta_{\textrm{NDCG}} \cdot \log \sigma (\FF(\cui, \phivi) - \FF(c_{u_j}, \phi_{v_j}))]
\end{equation}
where $\sigma$ refers to the sigmoid function and $\Delta_{\textrm{NDCG}}$ is the changes of a ranking metric, \ie, Normalized Discounted Cumulative Gain (NDCG)~\cite{jarvelin2002cumulated} in particular, after switching the ranking position of $i$ and $j$, so that ranking failures on top items are emphasized.

\highlight{Triplet regularization.} As the tuning experience triplets to train the ranker may not be ``plenty'' enough, it is critical to have a proper regularization. Similar to previous works~\cite{mittal2020hyperstar,jomaa2021dataset2vec}, we use a triplet loss with margin~\cite{schultz2004learning} to minimize the distance to a batch from the same dataset minus the distance to another dataset. Formally, it is defined as, 
\begin{equation}
\Loss_{\textrm{sim}}(G)= \max \left( \| \phivi - \widetilde{\phivi} \|^2 - \| \phivi - \phi_{v_j} \|^2 + \alpha, 0 \right)
\end{equation}
where $v_i \ne v_j$. $\phivi$ and $\widetilde{\phivi}$ are the meta-feature of the same dataset but extracted with different representative image samples, $\phi_{v_j}$ and $\phivi$ are different datasets and $\alpha$ is a hyper-parameter controlling the margin. Note that this loss is only applicable when $v_i = \tau$ or $v_j = \tau$. Otherwise, $G$ will not receive any gradients. 


\highlight{Synaptic Intelligence.} Though we used joint training to incorporate the experience from both past and present, as discussed previously, only the configuration encoder is optimized. Although the whole performance ranker benefits from those updates, the meta-feature extractor could still experience forgetting issue, which potentially degrades the performance. Because the meta-feature extractor is federatedly trained on the client side, we do not have access to their raw data. 
Thus, we adopt another continual learning technique, Synaptic Intelligence (SI)~\cite{zenke2017continual}, which is a regularization loss that nicely fits our framework. To this end, for any dataset $d_t$, we first calculate $\omega^{(t)}$, which is a per-parameter contribution to the change of loss:
\begin{equation}
\omega^{(t)}=- \sum_{i=1}^{N_{\text {iters }}}\left(\theta^{(t,i)}-\theta^{(t,i-1)}\right) \odot \frac{\delta \mathcal{L}_{\text {unreg}}^{(t,i)}}{\delta \theta^{(t,i)}}
\end{equation}
where $N_{\text {iters}}$ denotes the number of iterations. $\theta^{(t,i)}$ is the parameters after the $i$-th iteration of training on dataset $d_t$. $\mathcal{L}_{\text {unreg}}^{(t,i)}$ represents the loss without this regularization term at the $i$-th iteration. $\odot$ means element-wise product. We sum $\omega$ over the datasets to get $\Omega^{(\tau-1)}$, which the importance of every parameter in the first $\tau - 1$ datasets ($\tau$ is current time):
\begin{equation}
\Omega_{i}^{(\tau-1)}=\sum_{t=1}^{\tau-1} \frac{\omega_{i}^{(t)}}{\left(  \theta_i^{(t)} - \theta_i^{(t-1)} \right)^{2}+\xi}
\end{equation}
where $\theta_i$ is the $i$-th parameter in $\theta$. $\xi$ is a small dampening term to prevent dividing by zero, which we set to 0.1. Then, the regularization loss is given by,
\begin{equation}
\Loss_{\textrm{reg}}(F) = \sum_{i=1}^{|\theta|} \Omega_i^{(\tau-1)} (\theta_i - \theta_i^{(\tau-1)})^2
\end{equation}


Overall, the loss for the performance predictor is,
\begin{equation}
\label{eq:loss-total}
\Loss_{\text{total}}(F) = \Loss_{\text{rank}}(F)+\lambda_{\text{sim}}\Loss_{\text{sim}}(G) + \lambda_{\text{reg}} \Loss_{\text{reg}}(F)
\end{equation}
$\lambda_{\text{sim}}$ and $\lambda_{\text{reg}}$ are hyper-parameters controlling the weight of regularizations.

\section{Experiments}



\newcommand{\UP}{($\uparrow$)}
\newcommand{\DOWN}{($\downarrow$)}


\subsection{Experiment setup}
\label{sec:experiment-setup}



\highlight{Face detector training.} We use RetinaFace~\cite{deng2019retinaface} with MobileNet-V2~\cite{sandler2019mobilenetv2} backbone, pretrained on WIDER-Face~\cite{yang2016wider}. To train on a new dataset, we inherit weights pretrained on WIDER-Face, and fine-tune it on the target dataset. If the model used a different architecture from the pre-trained model, we perform a network adaptation with parameter remapping~\cite{fang2019fast}. We adopt ``Reduce LR on Plateau'' learning rate scheduler to ensure convergence. For evaluation, we follow \cite{CenterFace,deng2019retinaface,Linzaer} to rescale the shorter side of images to 720 pixels. 
We use Average-Precision at IoU 0.5 (AP@50) as our evaluation metric. A more detailed training setup is provided in Appendix B.1. 

\highlight{Datasets.} We evaluate \algname{} on 12 publicly available face datasets: AFLW~\cite{koestinger11a}, Anime~\cite{qhgz2013}, FaceMask~\cite{wobot2020facemask}, FDDB~\cite{fddbTech}, FDDB-360~\cite{fu2019fddb360}, MAFA~\cite{Ge_2017_CVPR}, Pascal VOC~\cite{Everingham15}, UFDD~\cite{nada2018pushing}, UMDAA-02~\cite{mahbub2017partial}, WIDER-Face~\cite{yang2016wider}, WIDER-360~\cite{fu2019datasets}, WIKI~\cite{Rothe-ICCVW-2015}. Details on data cleaning and train/val/test split can be found in Appendix B.2.
We split the datasets into server side and client side, where
WIDER-Face is considered always available at the server side. The rest of the datasets are treated as customer data which is not directly visible to the central server. We shuffle the order of the 11 customer datasets before every experiment, so as to simulate the scenario of customers' data coming by in an arbitrary order.

\highlight{Search space.} We define two search spaces for our evaluation. (i) \textbf{HPO space} (\ie hyper-parameter search space) tunes 6 different dimensions and contains 216 combinations of hyper-parameters, spanning from generic hyperparameters (\eg, learning rate) to domain-specific ones (\eg, IoU threshold). (ii) \textbf{NAS space} tunes the backbone architecture, as backbones are found to be essential to the detection performance~\cite{chen2019detnas}. Our backbone search space is a MobileNetV2-like space proposed by \cite{cai2018proxylessnas}. We limit the FLOPs to be less than 730M under 360P resolution to avoid selecting extra large models. The size of whole search space is $2.44 \cdot 10^{15}$. Details available in Appendix B.3.

\highlight{Dataset augmentation.} With only WIDER-Face available at the beginning, it is difficult to obtain a meaningful meta-feature extractor. Also, the training of meta-feature transformation module relies on a diversity of datasets so that it does not overfit to a particular representation. Following \cite{xue2019transferable,xu2021nasoa}, we extract subsets from WIDER-Face to create a variety of datasets on the server side. We intentionally manipulate the distribution of each subset to increase diversity in domains. This is done by clustering features generated by different deep learning models. We end up creating 1418 datasets to form $\Doff$. Refer to Appendix B.4 for details.

\highlight{Performance ranker.}
For each of HPO and NAS space, we sample 3,000 pairs of configurations and datasets (from $\Doff$), and get their corresponding performance. This forms a meta-dataset which we use to ``\emph{warm-up}'' the ranker before online datasets are received. 
During online stage, for each dataset, we randomly sample 200 configurations (that almost exhausts the space for HPO), and get $B$ configurations according to the prediction scores. 
An exploration-exploitation strategy~\cite{dudziak2020brp} is adopted to prevent the selected configurations becoming too homogeneous. 
Following \cite{mittal2020hyperstar}, we pre-built a performance \emph{benchmark} (\ie, lookup table), so that we do not have to go through the computationally-expensive step of detector training in every experiment.
Budget, \ie, the number of trials for each dataset, is set to 4 if not otherwise specified. Other hyper-parameters can be found in Appendix B.5.

\subsection{Performance of \algname{}}

\highlight{Baselines.} We compare our method with the following baselines:
(i) \textbf{Random search}: Randomly selecting certain number of configurations, without any knowledge over dataset or search space.
(ii) \textbf{Best on WIDER}: Finding the top configurations on WIDER-Face  (by grid search on HPO space and performance predictor on NAS space) and applying them on new datasets.
(iii) \textbf{Tr-AutoML}~\cite{xue2019transferable}: A Markov-analysis based method. It can be applied to our scenario because it is a transferable AutoML algorithm that is designed for an online setting similar to ours.
(iv) \textbf{HyperSTAR}~\cite{mittal2020hyperstar}: The framework is optimized in an end-to-end manner. It adopts a frozen ResNet50~\cite{he2015deep} as meta-feature extractor, and thus can be adapted to our scenario. We fine-tune its predictor on all historic experience before serving new datasets. 
(v) \textbf{SCoT}~\cite{bardenet2013collaborative}: A bayesian optimization framework whose surrogate model learns to predict performance conditioned jointly on descriptive statistics of datasets and configurations. 

\highlight{Evaluation metrics.} The metrics used in our evaluation are:
(i) \textbf{$\Delta$AP}: The performance gain of the proposed search algorithm compared to random search under the same budget. The gains are \emph{summed} across datasets. The higher the better \UP.  
 (ii) \textbf{Rank}: The rank of the best found configuration in the search space. The rank is normalized to 0$\sim$100\% for easy comparison across different search spaces, then averaged over datasets. The lower the better \DOWN.

\begin{table}[t]
\centering
\resizebox{\columnwidth}{!}
{
\begin{tabular}{l|cc|cc}
\hline
\multirow{2}{*}{Method} & \multicolumn{2}{c|}{HPO space} & \multicolumn{2}{c}{NAS space} \\
\cline{2-5}
  & $\Delta$AP \UP & Rank \DOWN 
  & $\Delta$AP \UP & Rank \DOWN \\ 
\hline
      Random search &          0.00 &          20.09 &          0.00 &         20.10 \\
      Best on WIDER &          1.33 &          25.12 &          2.04 &         10.97 \\
          Tr-AutoML~\cite{xue2019transferable} &         -0.18 &          22.18 &         -0.52 &         25.29 \\
          HyperSTAR~\cite{mittal2020hyperstar} &         -0.47 &          22.00 &         -0.14 &         21.09 \\
              SCoT~\cite{bardenet2013collaborative} &         -0.10 &          20.83 &         -0.28 &         21.30 \\
\hline
\algname{} (ResNet) &         -0.23 &          17.41 &          1.54 &         11.97 \\
 \algname{} (Statistics) &          0.08 &          17.59 &          1.58 &         12.06 \\
  \algname{} (MSE) &         -0.05 &          20.93 &          0.21 &         18.34 \\
\hline
         \algname{} & \textbf{1.67} & \textbf{13.16} & \textbf{2.39} & \textbf{7.78} \\
\hline
\end{tabular}
}
\caption{Comparison of end-to-end performance with baselines and several variants. \textbf{\algname{} (ResNet)}: using ResNet50 as meta-feature extractor. \textbf{\algname{} (Statistics)}: using descriptive statistics as meta-features. \textbf{\algname{} (MSE)}: using MSE as the primary loss.}
\vspace{-2em}
\label{tab:end-to-end}
\end{table}

\highlight{Results.} We show the results in \autoref{tab:end-to-end}. We evaluate each setting with 20 different seeds and report the average. The standard deviations along with more detailed results on each dataset can be found in Appendix C.

We firstly point out that we cannot rely on the tuning experience of one single dataset. The top configurations on WIDER-Face rank 25.12\% and 10.97\% in average for HPO and NAS search space respectively, indicating that no golden configuration works on all datasets. Although on some datasets, the best configuration on WIDER-Face can be impressive (\eg, WIDER-360, HPO, +2.5\% $\Delta$AP), for some datasets they can be almost at the bottom (\eg, ANIME, HPO, worse than 85.6\% of the search space).

Surprisingly, in our scenario, the sophisticated approaches (\eg, HyperSTAR) perform even worse than a simple random search baseline. We think this could be attributed to three reasons. (i) The meta-feature they proposed to measure the similarity of datasets does not align with the face detection scenario. (ii) The loss functions they used (\eg, MSE) make the optimization difficult. (iii) For Tr-AutoML, the design of sharing the ``top-1'' configuration among datasets is too arbitrary and does not sufficiently exploit the budget. To verify our hypothesis, we create variants of \algname{} by replacing the meta-feature extractor and loss function with those used in baselines. The results are shown in \autoref{tab:end-to-end}, which are significantly worse than \algname{}, indicating that the meta-feature extractor and loss function tailored for our scenario are the indispensable components to make the algorithm work. Among the three variants, the MSE version has the worst performance, implying that ranking loss is the most crucial to our results. 

\begin{figure}[t]
\centering
\includegraphics[width=\columnwidth]{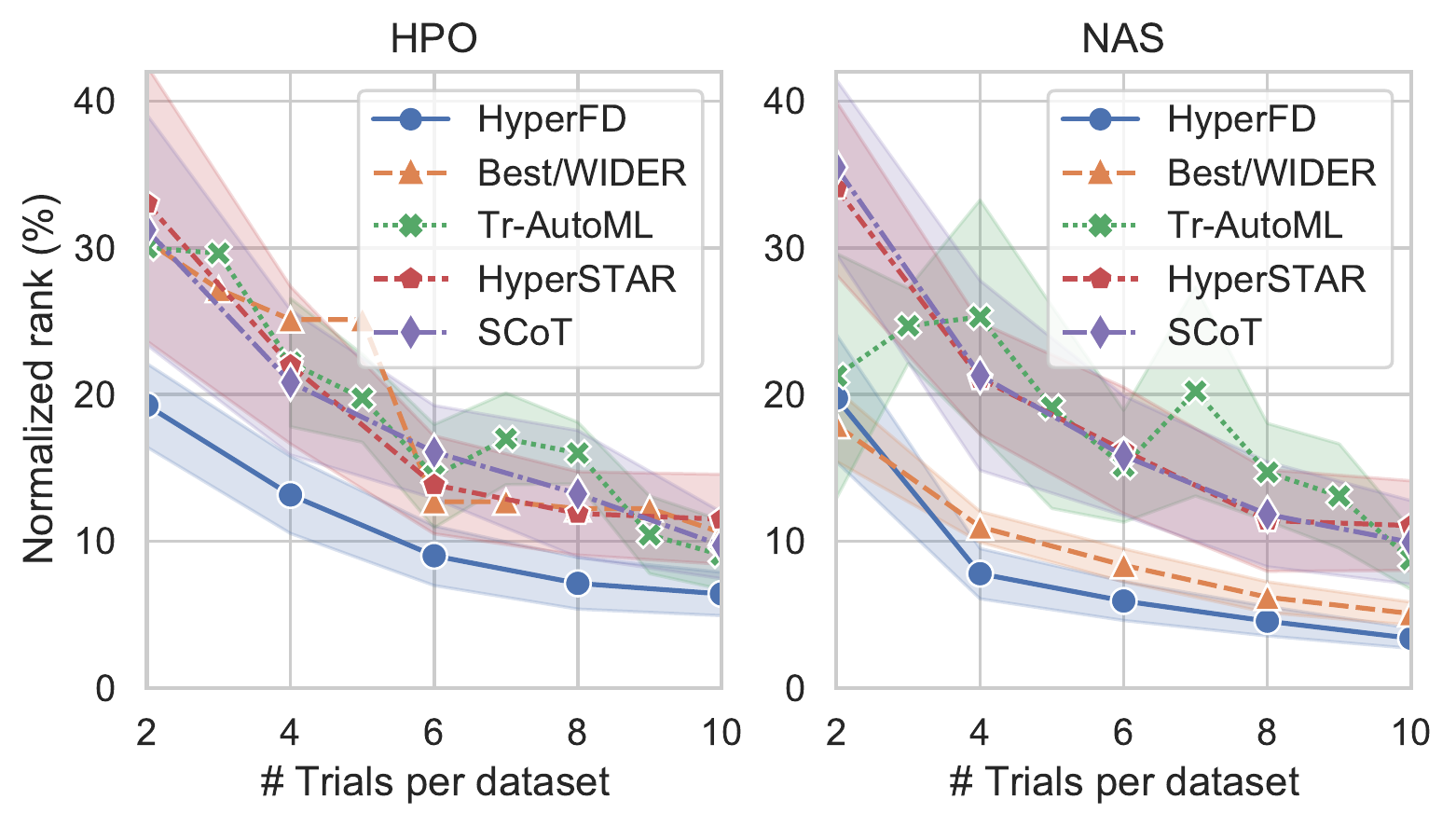}
\caption{Performance (rank normalized to 0--100\%) of \algname{} compared to baselines when assigned with different budgets. The shaded region is the standard deviation for 20 runs.}
\label{fig:budget-rank}
\vspace{-2em}
\end{figure}

\algname{} remarkably outperforms baselines on all metrics, especially in terms of ``rank''. 
It is noteworthy that the improvement is particularly significant on the HPO search space. By contrast, on the NAS search space, most of the algorithms enjoy a better rank. We argue that NAS is an easier search space for AutoML, mainly because the datasets share more preferences for architectures than for hyper-parameters. The findings in \S~\ref{sec:analysis} also echo such claim.

Finally, we verify the effectiveness of \algname{} by varying the budget, \ie, number of trials to run for each dataset, between 2 and 10. We then report the averaged rank for each method under different budgets. The results are shown in \autoref{fig:budget-rank}. Remarkably, \algname{} almost consistently performs best under all variants of budgets.  

\subsection{Ablations}

\highlight{Continual learning.} We first evaluate the effectiveness of the proposed continual learning strategy. Specifically, we conduct several ablation experiments including: (i) Only leverage the tuning experience from the latest dataset. (ii) Freeze the meat-feature extractor after warm-up. (iii) Freeze the whole performance ranker after warm-up. The results are in \autoref{tab:ablation-continual}. Thanks to the design of our performance ranker and the warm-up process, even if the whole ranker is frozen, the performance still looks fairly good. 
Nevertheless, using continual learning techniques can make the framework perform even better.

\begin{table}[t]
\centering
\resizebox{\columnwidth}{!}
{
\begin{tabular}{l|cc|cc}
\hline
\multirow{2}{*}{Method} & \multicolumn{2}{c|}{HPO} & \multicolumn{2}{c}{NAS} \\
\cline{2-5}
& $\Delta$AP \UP & Rank \DOWN & $\Delta$AP \UP & Rank \DOWN \\
\hline
     Latest data only &          1.12 &          15.99 &          2.03 &          9.55 \\
Freeze meta-extractor &          1.22 &          15.28 &          2.19 &          9.41 \\
  Freeze whole ranker &          1.32 &          16.85 &          2.26 &          8.60 \\
    \algname{} (full) & \textbf{1.67} & \textbf{13.16} & \textbf{2.39} & \textbf{7.78} \\
\hline
\end{tabular}
}
\caption{Ablations to justify the necessity of continual learning.}
\label{tab:ablation-continual}
\vspace{-1em}
\end{table}

\highlight{Transformation module.} The proposed transformation module is a critical component of \algname{}. Equipped with the transformation module, \algname{} can benefit from the historical experience without sacrificing any privacy, because the transformation module only needs abstract dataset-level meta-features rather than raw images. From the results in \autoref{tab:ablation-transformation}, we can learn that the performance drops significantly without the transformation module, which means our design enables a more effective usage of the historical experience and alleviates the knowledge forgetting. Furthermore, we also compare \algname{} with the oracle baseline which neglects the privacy concerns and gathers all data at the central server. Remarkably, our transformation obtains comparable performance to the oracle baseline, which again proves the effectiveness.


\begin{table}[t]
\centering
\resizebox{\columnwidth}{!}
{
\begin{tabular}{l|cc|cc}
\hline
\multirow{2}{*}{Method} & \multicolumn{2}{c|}{HPO} & \multicolumn{2}{c}{NAS} \\
\cline{2-5}
& $\Delta$AP \UP & Rank \DOWN & $\Delta$AP \UP & Rank \DOWN \\
\hline
No transformation &         1.28 &        15.65 &         2.16 &         9.67 \\
           Oracle &         1.84 &        13.79 &         2.37 &         7.73 \\
\algname{} (full) &         1.67 &        13.16 &         2.39 &         7.78 \\
\hline
\end{tabular}
}
\caption{Ablations on transformation module.}
\label{tab:ablation-transformation}
\vspace{-1em}
\end{table}

\highlight{Loss functions.} The effectiveness of our loss functions are demonstrated in \autoref{tab:ablation-loss}. There are three components in our loss function: a ranking loss and two regularization losses (triplet loss and SI loss). The ranking loss is essential and cannot be disabled, thus we turn the other two regularization losses on and off for comparison. On average, the triplet loss and SI loss contributes 3.81\% and 1.66\% to the rank across different search spaces.  

\begin{table}[t]
\centering
\resizebox{\columnwidth}{!}
{
\begin{tabular}{ccc|cc|cc}
\hline
\multirow{2}{*}{Ranking} & \multirow{2}{*}{Triplet} & \multirow{2}{*}{SI} & \multicolumn{2}{c|}{HPO} & \multicolumn{2}{c}{NAS} \\
\cline{4-7}
& & & $\Delta$AP \UP & Rank \DOWN & $\Delta$AP \UP & Rank \DOWN \\
\hline
             $\checkmark$ & & $\checkmark$ &          0.50 &          16.49 &          1.60 &         12.06 \\
            $\checkmark$ & $\checkmark$ &  &          1.38 &          15.40 &          2.33 &          8.86 \\
$\checkmark$ & $\checkmark$ & $\checkmark$ & \textbf{1.67} & \textbf{13.16} & \textbf{2.39} & \textbf{7.78} \\
\hline
\end{tabular}
}
\caption{Ablations on loss functions.}
\label{tab:ablation-loss}
\vspace{-1.5em}
\end{table}

\highlight{Designs of meta-feature extractor.} Apart from variants of meta-feature extractors in \autoref{tab:end-to-end}, we further experiment with various versions, 
where we disable several key components in our meta-feature extractor. For a more comprehensive comparison, we further show the validation NDCG~\cite{jarvelin2002cumulated}, \ie a ranking metric used to assess the warm-up quality. \autoref{tab:ablation-meta-feature} empirically proves the effectiveness of our meta-feature extractor.
Of the components in our extractor, the ablation of self-attention has the largest effect, with the worst NDCG(val) on both search spaces.

\begin{table}[t]
\centering
\resizebox{\columnwidth}{!}
{
\begin{tabular}{l|cc|cc}
\hline
\multirow{2}{*}{Method} & \multicolumn{2}{c|}{HPO} & \multicolumn{2}{c}{NAS} \\
\cline{2-5}
& NDCG(val) \UP & Rank \DOWN & NDCG(val) \UP & Rank \DOWN \\
\hline
Anchor-level: w/o labels &          0.930 &          14.13 &          0.895 &         11.13 \\
   Image-level: w/o GWAP &          0.932 &          14.22 &          0.901 &         10.41 \\
Image-level: w/o pyramid &          0.927 &          16.92 &          0.898 &         12.98 \\
  Image-level: w/o statistics &          0.922 &          15.42 &          0.899 &         11.92 \\
Dataset-level: w/o attn. &          0.919 &          15.09 &          0.886 &         12.09 \\
\hline
\algname{} (full) & \textbf{0.933} & \textbf{13.16} & \textbf{0.903} & \textbf{7.78} \\
\hline
\end{tabular}
}
\caption{Ablations on meta-feature extractor. \textbf{w/o labels}: do not attach the bounding boxes information to feature maps. \textbf{w/o GWAP}: use average pooling rather than GWAP. \textbf{w/o pyramid}: use features from the last stage only. \textbf{w/o statistics}: no dataset-level descriptive statistics attached to the feature of each image. \textbf{w/o attn.}: remove the self-attention encoder layer.}
\label{tab:ablation-meta-feature}
\end{table}

\highlight{Warm-up and augmentation.} Finally, we ablated our warm-up that is performed before the algorithm receives online datasets. We compare our method with the version without warm-up at all, and the version with warm-up but no augmented datasets. Results can be found in \autoref{tab:ablation-warm-up}, where we see a performance drop without warm-up or augmentation. We argue that without proper warm-up, both meta-feature extractor and configuration encoder are prone to overfit the few samples (\ie, trials) on each task. Moreover, meta-feature transformation module can hardly learn anything useful with only WIDER-Face dataset in hand. 

\begin{table}[t]
\centering
\resizebox{\columnwidth}{!}
{
\begin{tabular}{l|cc|cc}
\hline
\multirow{2}{*}{Method} & \multicolumn{2}{c|}{HPO} & \multicolumn{2}{c}{NAS} \\
\cline{2-5}
& $\Delta$AP \UP & Rank \DOWN & $\Delta$AP \UP & Rank \DOWN \\
\hline
      w/o warm-up &          0.50 &          16.28 &          1.38 &         12.62 \\
 w/o augmentation &          0.96 &          17.55 &          1.43 &         15.23 \\
\algname{} (full) & \textbf{1.67} & \textbf{13.16} & \textbf{2.39} & \textbf{7.78} \\
\hline
\end{tabular}
}
\caption{Ablations for warm-up and augmentation.}
\label{tab:ablation-warm-up}
\vspace{-1em}
\end{table}

\subsection{Analysis}
\label{sec:analysis}

\begin{figure}[t]
\centering
\includegraphics[width=\linewidth]{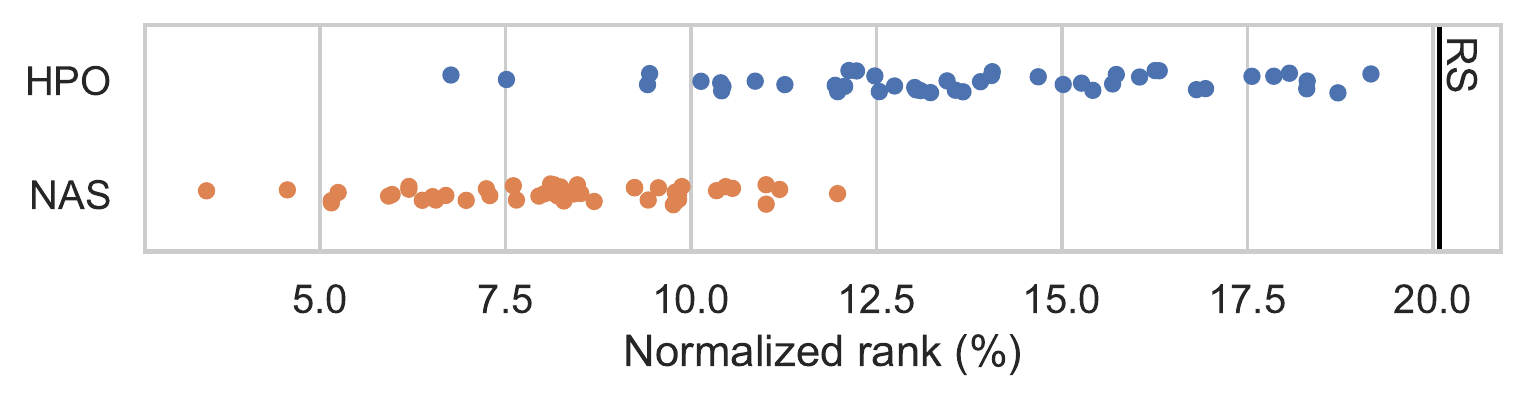}
\vspace{-2em}
\caption{Distribution of normalized ranks of 50 experiments. RS: Random search.}
\label{fig:distribution-rank}
\vspace{-1.5em}
\end{figure}

\highlight{Robustness to task arriving order.} The proposed \algname{} is robust to the order in which the tasks arrive. In \autoref{fig:distribution-rank}, we run 50 experiments with different arriving order for each search space, and report their distribution. Even the worst case is still much better than the average case of random search. Such robustness to order ensures the fairness across different tasks~\cite{yoon2020scalable}, which is important for a reliable platform.

\begin{figure}[t]
\centering
\vspace{-1em}
\begin{subfigure}{0.49\columnwidth}
\includegraphics[width=\columnwidth]{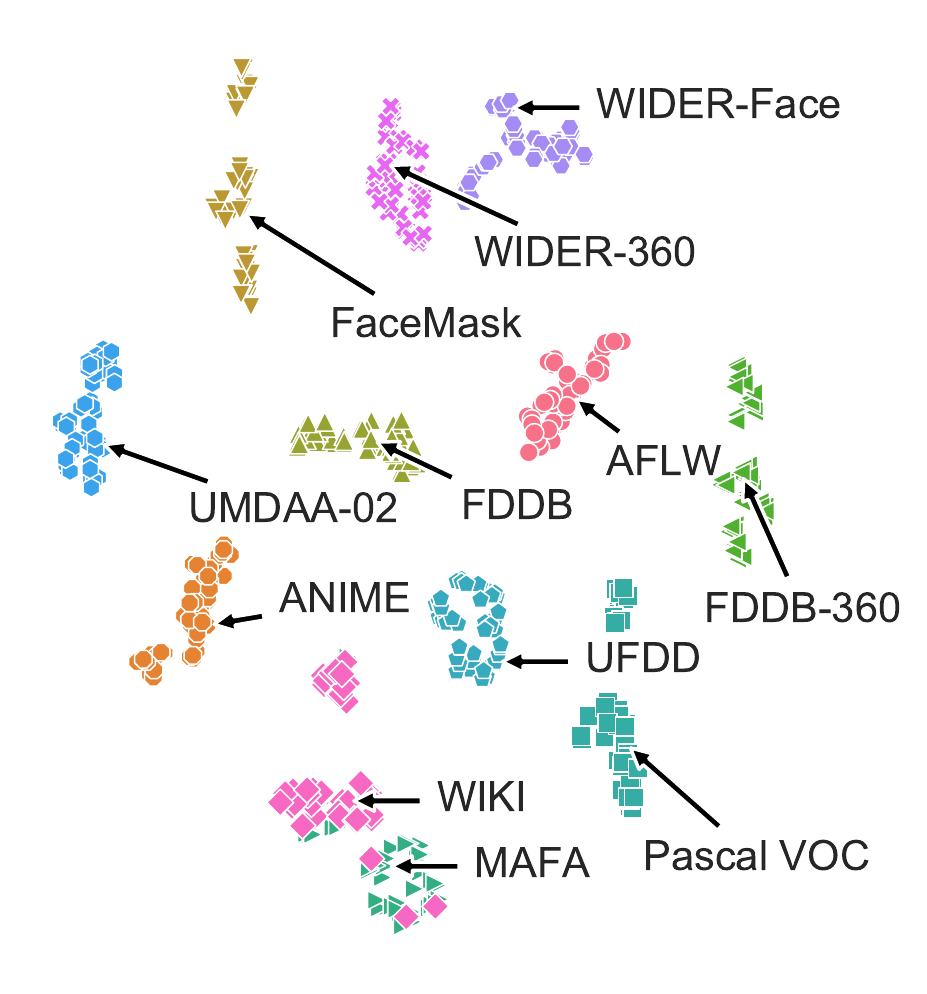}
\end{subfigure}
\begin{subfigure}{0.49\columnwidth}
\includegraphics[width=\columnwidth]{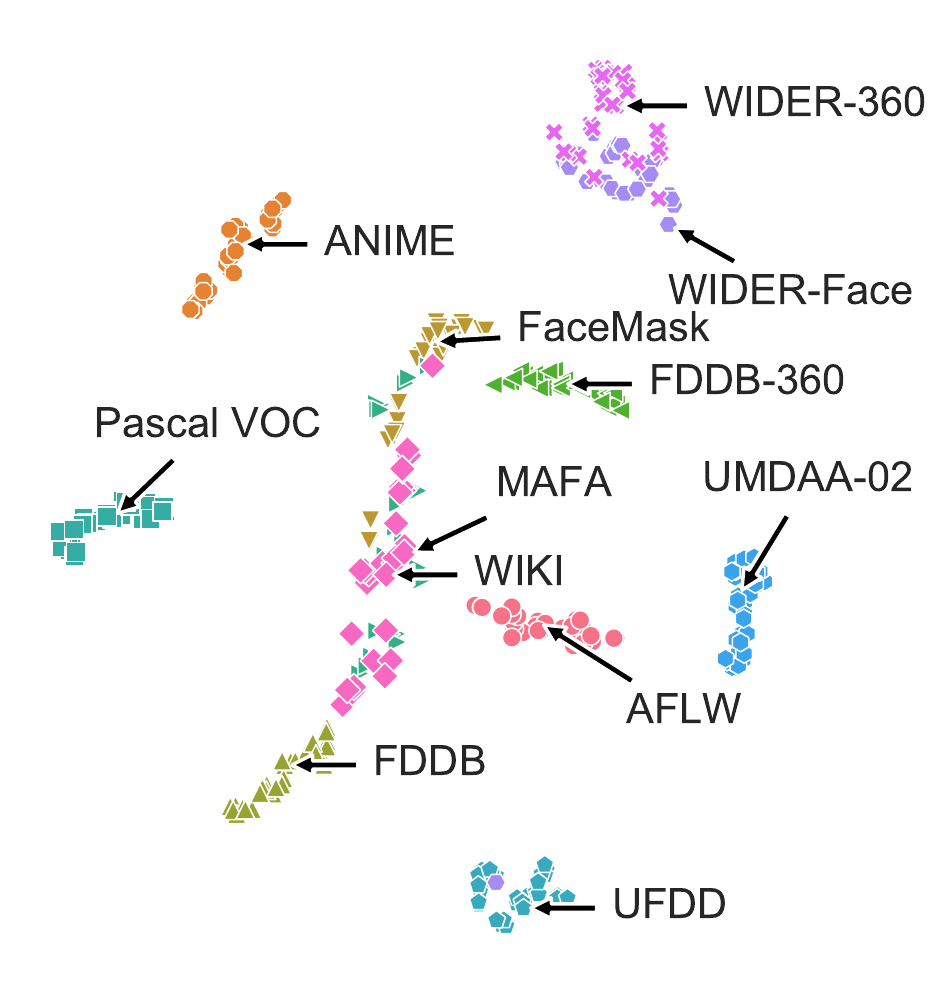}
\end{subfigure}
\vspace{-1em}
\caption{t-SNE visualization of the feature representations after offline warm-up. Each color represents a dataset. Multiple points correspond to multiple batches of images. \textbf{(left)} Trained on HPO search space. \textbf{(right)} Trained on NAS search space. (Better viewed in color)}
\label{fig:tsne-anime}
\vspace{-1.5em}
\end{figure}

\highlight{Meta-feature visualization.} We visualize representation of meta-features via t-SNE~\cite{JMLR:v9:vandermaaten08a}. For each dataset, we extract meta-feature 30 times, each of which is extracted from a randomly sampled batch of images from the dataset. Results are shown in \autoref{fig:tsne-anime}. After warmed up, our meta-feature extractor can already successfully distinguish most of the datasets without further training the extractor on those datasets. There are two interesting findings. First, the datasets which are similar to each other (\eg, WIDER-Face and WIDER-360) are also located close to each other in the figure. Second, the datasets tend to be more mingled among each other on NAS search space than on HPO search space, which means the preference on architectures is easier to be transferred among datasets than the preference on hyper-parameters. This also justifies the necessity to optimize the meta-feature extractor in an end-to-end manner.

\section{Discussions and conclusions}


From a platform perspective, this paper studied online AutoML for domain-specific face detection, \ie how to continuously improve the AutoML algorithm and learn from a sequence of training tasks while protecting the privacy of sensitive face detection data. Various techniques are proposed and extensive experiments show their effectiveness. 
Admittedly, this paper did not include incorporation with multi-fidelity techniques (\eg, BOHB~\cite{falkner2018bohb}), and did not attempt to generalize our approach to scenarios other than face detection. We leave them in future works.

\highlight{Broader impacts.} The sensitivity of face datasets has been long noticed, and there is no need to emphasize more the importance of protect their privacy. However, if customers use their own datasets isolatedly, such over-protection will put resources in waste, especially when AutoML is used. By introducing \algname{}, we reach a balance between efficiency and privacy, and achieve the best of both worlds. 


\section*{Acknowledgements}

We thank anonymous reviewers for their valuable feedbacks. We also thank Chengmin Chi (STCA) and Xiaotian Gao (MSRA) for their suggestions.

{\small
\bibliographystyle{ieee_fullname}
\bibliography{egbib}
}

\clearpage
\appendix

\section{Pseudo-code for \algname{}}

We summarize the flow of the full algorithm as follows.

\begin{algorithm}
	\renewcommand{\algorithmicrequire}{\textbf{Input:}}
	\renewcommand{\algorithmicensure}{\textbf{Output:}}
	\caption{\algname{}}
	\label{alg:hyperfd}
	\begin{algorithmic}
		\REQUIRE Search space of configurations $\mathcal{C}$. Offline prepared datasets $\mathcal{D}_{\textrm{offline}}$. A sequence of online datasets $\mathcal{D} = \{d_1, d_2, \ldots, d_N\}$. Detector training and evaluation pipeline $\textrm{AP}(c, d)$. Budget for each task $B$. Batches to update for each task $N_{\textrm{iters}}$. Iterations of updates of transformation module $N_{\textrm{trans}}$. Learning rates $\eta_{z}$ and $\eta_{\theta}$.
		\ENSURE Best configurations $\{c_1^*, c_2^*, \ldots, c_N^*\}$.

        \STATE {$\triangleright$ \textit{Warm-up}}
        \STATE $S_{\textrm{offline}} \leftarrow \{(c, d, \textrm{AP}(c, d)) ~ | ~ c \in \mathcal{C}, d \in \mathcal{D}_{\textrm{offline}} \}$.
        \STATE $\theta \leftarrow \argmin_{\theta}\mathcal{L}_{\textrm{total}} (F; S_{\textrm{offline}})$.
        \STATE $S_{\textrm{past}} \leftarrow \{\}$.
        \STATE $\theta^{(0)} \leftarrow \theta$.
        
        \STATE{}

        \FOR {$t = 1, 2, \ldots, N$}
    		\STATE {$\triangleright$ \textit{Inference}}
            \STATE {Get top-$B$ configs $C_t = \{c_1, c_2, \ldots, c_B\}$ with $F$.}
            \STATE {$c_t^* \leftarrow \argmax_i \textrm{AP}(C_{t,i}, d_t)$. \quad $\triangleright$ \textit{Costly step}}
    		
    		\STATE{}
    		
    		\STATE {$\triangleright$ \textit{Training}}
    		
    		\STATE {$S \leftarrow \{(c, d_t, \textrm{AP}(c, d_t) ~ | ~ c \in C_t \}$}
    		\FOR {$i = 1, 2, \ldots, N_{\text{iters}}$}
        		\STATE {$\triangleright$ \textit{Training of transformation module}}
    		    \FOR {$k = 1, 2, \ldots, N_{\textrm{trans}}$}
    		        \FOR {$u = 1, 2, \ldots, t - 1$}
                        \STATE{$\textbf{Z}^{(u)} \leftarrow \textbf{Z}^{(u)} - \eta_{z} \nabla_{\textbf{Z}^{(u)}} \mathcal{L}_{\textrm{trans}} (\textbf{Z}^{(u)}; \mathcal{D}_{\textrm{offline}})$.}
    		        \ENDFOR
    		    \ENDFOR
    		    \STATE {$\triangleright$ \textit{Training of performance ranker}}
    		    \STATE{$\theta \leftarrow \theta - \eta_{\theta} \nabla_{\theta} \mathcal{L}_{\textrm{total}} (F; S \bigcup S_{\textrm{past}})$.}
    		\ENDFOR
    		
    		\STATE{}

    		\STATE $\theta^{(i)} \leftarrow \theta$.
    		\STATE {$S_{\textrm{past}} \leftarrow S_{\textrm{past}} \bigcup \{(c, \phi_t, \textrm{AP}(c, d_t) ~ | ~ c \in C_t \}$}.
		\ENDFOR
	\end{algorithmic} 
\end{algorithm}

\section{Experiment setups}

\subsection{Detector training}

We use RetinaFace~\cite{deng2019retinaface} with MobileNet-V2~\cite{sandler2019mobilenetv2} (channels $\times 0.5$ variant). The channels of FPN and context modules are set to 80. The backbone is firstly pre-trained on ImageNet~\cite{deng2009imagenet}, and then the full detector is further pre-trained on WIDER-Face~\cite{yang2016wider} to boost the performance of the model, especially on small datasets like PASCAL~\cite{Everingham15}. For training, we set batch size to 32, image size to $640 \times 640$ (after crop), and uses LeakyReLU as activation functions. We perform a validation per 10 epochs of training, and we adopt a ``Reduce LR on Plateau'' policy that decays the learning rate by 10 when metric on validation set stagnates in the 5 past evaluations. The maximum epochs of training is 1000, but we stop the training when the validation performance no longer increases for 8 times. For evaluation, we follow \cite{CenterFace,deng2019retinaface,Linzaer} to rescale the shorter side of images to 720 pixels, ignore bounding boxes smaller than 36 pixels, and perform Non-maximum Suppression (NMS) on overlap predictions with a 0.4 IoU threshold. We select the best model in history and evaluate it on test dataset. We use Average-Precision at IoU 0.5 (AP@50) as our evaluation metric, with the evaluation scheme implemented in MMDetection~\cite{chen2019mmdetection}.

\subsection{Detection datasets}

\begin{figure*}[t]
\centering
\includegraphics[width=\textwidth]{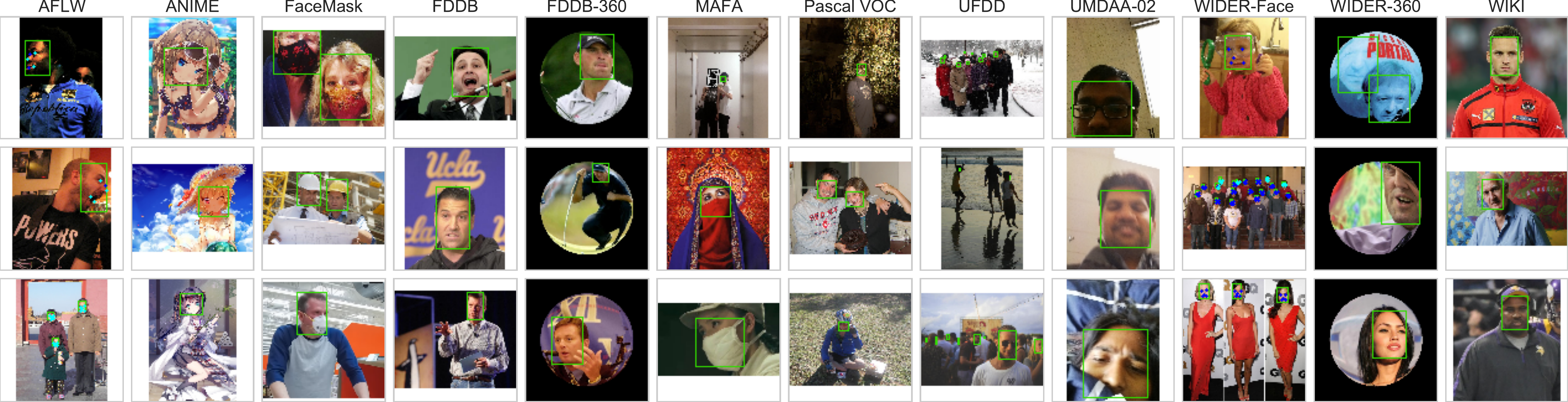}
\caption{Preview of 12 datasets. We show 3 images per dataset.}
\label{fig:fdd24-preview}
\end{figure*}

We gather 12 public datasets for our evaluation, which includes AFLW~\cite{koestinger11a}, Anime~\cite{qhgz2013}, FaceMask~\cite{wobot2020facemask}, FDDB~\cite{fddbTech}, FDDB-360~\cite{fu2019fddb360}, MAFA~\cite{Ge_2017_CVPR}, Pascal VOC~\cite{Everingham15}, UFDD~\cite{nada2018pushing}, UMDAA-02~\cite{mahbub2017partial}, WIDER-Face~\cite{yang2016wider}, WIDER-360~\cite{fu2019datasets}, WIKI~\cite{Rothe-ICCVW-2015}. The datasets span multiple categories, including faces with masks, anime faces, faces from fisheye cameras, selfies from cellphones and etc. A preview of the datasets is shown in \autoref{fig:fdd24-preview}. Apart from cleanup for illegal bounding boxes and corrupted images (particularly on WIKI, WIDER-Face, and UMDAA-02), the quality of annotations on AFLW and WIKI is particularly low, as they are designed for other vision tasks (\eg facial landmarks and face recognition). Thus, we only consider the original ground truth bounding boxes as a reference and use RetinaFace~\cite{deng2019retinaface} to ensure the quality of weak labels with bipartite matching towards old labels. Specifically, we replace the original ground truth labels with the weak labels if IoU threshold $> 0.4$ by bipartite matching. For unmatched labels, we mark the confidence $> 0.95$ as positive. For hard cases that detectors have little confidence, we manually checked each of them. We also unify the format of facial landmarks to 5 points (\ie eyes, noses, mouths), preserving the landmark annotations in AFLW (downsize the 19 landmarks) and WIDER-Face. Finally, we split each dataset into three splits: train, val and test, in the ratio of 6:1:3. This split is fixed and we will use it in all our following experiments. The overall statistics of all processed datasets are provided in \autoref{tab:fdd24-statistics}.

\begin{table}[t]
\centering
\begin{adjustbox}{width=\linewidth}
\begin{tabular}{cccccc}
\toprule
\multirow{2}{*}{Dataset} &  \multicolumn{3}{c}{\# Images} &  \# Faces &  \# Faces \\
              & Train & Val & Test & per image &    w. landm.        \\
\midrule
AFLW~\cite{koestinger11a}              &  12477 &  2079 &   6239 &    1.5 &         79\% \\
Anime~\cite{qhgz2013}             &   3703 &   617 &   1851 &    1.2 &          None \\
FaceMask~\cite{wobot2020facemask}          &   2554 &   425 &   1278 &    1.7 &          None \\
FDDB~\cite{fddbTech}             &   1693 &   282 &    846 &    1.8 &          None \\
FDDB-360~\cite{fu2019fddb360}          &   4867 &   811 &   2433 &    1.6 &          None \\
MAFA~\cite{Ge_2017_CVPR}              &  23261 &  2585 &   4927 &    1.0 &          None \\
Pascal VOC~\cite{Everingham15}            &    511 &    85 &    255 &    1.9 &          None \\
UFDD~\cite{nada2018pushing}              &   1781 &   296 &    892 &    3.7 &          None \\
UMDAA-02~\cite{mahbub2017partial}           &  16934 &  1881 &   4708 &    1.0 &          None \\
WIDER-Face~\cite{yang2016wider}             &   9665 &  1600 &   4832 &   12.2 &         80\% \\
WIDER-360~\cite{fu2019datasets}         &  38321 &  6394 &  19145 &    7.3 &          None \\
WIKI~\cite{Rothe-ICCVW-2015}              &  20941 &  3490 &  10471 &    1.2 &          None \\
\bottomrule
\end{tabular}
\end{adjustbox}
\caption{Statistical information of all datasets.} 
\label{tab:fdd24-statistics}
\end{table}

\subsection{Search space}

The hyper-parameter search space (HPO space) is shown in \autoref{tab:search_space}.

The neural architecture search space (NAS space) is shown in \autoref{tab:nas_search_space}. The space is essentially a channels $\times 0.5$ variant of ProxylessNAS search space~\cite{cai2018proxylessnas}. For detection purposes, we extract feature maps from the end of stage 3, 5 and 7, with down-sampling by 8, 16 and 32 respectively. To benefit from pre-training, we use parameter-remapping~\cite{fang2019fast}, where we train the largest network in the search space first, and map the weights to the target network with a set of rules.

\begin{table}[t]
\centering
    \begin{tabular}{|l|c|}
    \hline
    Optimizer & \{ SGD, Adam \} \\
    Learning rate &  $\{ 3 \cdot 10^{-4}, 10^{-3}, 3 \cdot 10^{-2} \}$ \\
    Min crop ratio & $\{ 0.3,  0.55 \}$ \\
    IoU threshold & $\{ 0.4, 0.5, 0.6 \}$ \\
    Location loss weight & $\{2, 4, 8 \}$ \\
    Neg-pos samples ratio & $\{ 2, 7 \}$ \\
    \hline
    \end{tabular}
\caption{HPO space contains 216 combinations of hyper-parameters. For IoU threshold, we show the positive matching threshold in the table, while the negative threshold $= \textrm{pos} / 2 + 0.05$.}
\label{tab:search_space}
\end{table}

\begin{table}[t]
\centering
\resizebox{\columnwidth}{!}{
\begin{tabular}{cccccc}
\toprule
Stage & Depth & Expand ratio & Kernel size & Width & Stride \\
\midrule
1 & 1 & 1 & 3 & 8 & 1 \\
2 & 1--3 & $\{4,6\}$ & $\{3,5,7\}$ & 12 & 2 \\
3 & 1--3 & $\{4,6\}$ & $\{3,5,7\}$ & 16 & 2 \\
4 & 2--4 & $\{4,6\}$ & $\{3,5,7\}$ & 32 & 2 \\
5 & 3--4 & $\{4,6\}$ & $\{3,5,7\}$ & 48 & 1 \\
6 & 2--4 & $\{4,6\}$ & $\{3,5,7\}$ & 80 & 2 \\
7 & 1 & $\{4,6\}$ & $\{3,5,7\}$ & 160 & 1 \\
\bottomrule
\end{tabular}
}
\caption{Search space for neural architectures.}
\label{tab:nas_search_space}
\end{table}

\subsection{Dataset augmentation}

Dataset augmentation intends to create a large number of diverse datasets. To this end, we first extract features for each image using pre-trained models with diverse usages collected from model zoos (\eg, ONNX model zoo\footnote{\url{https://github.com/onnx/models}}). We use intermediate layers of features so that fine-grained distribution information is not lost. \autoref{tab:augmentation-models} shows a list of the used pre-trained models.

\begin{table}[t]
\centering
\resizebox{\columnwidth}{!}{
\begin{tabular}{lll}
\toprule
Task & Model & Layer \\
\midrule
Classification & EfficientNet~\cite{tan2019efficientnet} & Last hidden layer \\
Classification & IBN-Net~\cite{pan2018two} & Last hidden layer \\
Classification & Places365~\cite{gkallia2017keras_places365} & Last hidden layer \\
Detection & DETR~\cite{carion2020detr} & Transformer encoder \\
Detection & RetinaNet~\cite{lin2017focal} & FPN \\
Detection & CenterFace~\cite{CenterFace} & Before heads \\
Face recognition & ArcFace~\cite{deng2019arcface} $^\dag$ & Final layer \\
Face emotion & FerPlus~\cite{barsoum2016training} & Last hidden layer \\
Face age & VGG16-Age~\cite{lapuschkin2017understanding} & Final layer \\
Face gender & VGG16-Gender~\cite{lapuschkin2017understanding} & Last hidden layer \\
Label statistics $^\ddag$ & -- & -- \\
\bottomrule
\end{tabular}
}
\caption{List of pre-trained models used to generated features for augmentation. $\dag$: For ArcFace, we have two variants which feed the full image and the dominant face crop respectively. $\ddag$: Areas of bounding boxes on the image, described by \texttt{scipy.describe}.}
\label{tab:augmentation-models}
\end{table}

For each set of generated features, we run multiple combinations of clustering algorithms and configurations to get a diverse series of clustering results. The list is described in \autoref{tab:clustering-algos}. The implementations are with scikit-learn, and we throw away too small sub-datasets generated (less than 800 images) and too large subsets (less than 800 images are not selected). We respect the original train-validation-test split, and also remove subsets where the split becomes too deviated from balance. Finally, we have got 1418 sub-datasets generated from the original WIDER-Face dataset.

\begin{table}[t]
\centering
\resizebox{\columnwidth}{!}{
\begin{tabular}{lll}
\toprule
\# & Algorithm & Configuration \\
\midrule
1 & K-Means & MaxIter=2000, k-means++ init \\
2 & K-Means & MaxIter=2000, random init \\
3 & PCA + K-Means & \#components=20 \\
4 & PCA + t-SNE + K-Means & \#components=2 \\
5 & DBSCAN & eps=10, metric=l2 \\
6 & DBSCAN & eps=10, metric=cosine \\
7 & Agglomerative & \\
8 & Agglomerative & affinity=mahattan \\
9 & Birch~\cite{zhang1996birch} & \\
\bottomrule
\end{tabular}
}
\caption{List of clustering algorithms and configurations to group the images based on features. The ``number of clusters'' is iterated in 2, 3, 5 and 7 for each configuration. The interpretation of configurations corresponds to parameters in scikit-learn.}
\label{tab:clustering-algos}
\end{table}

\subsection{Performance ranker}

Important hyper-parameters for performance ranker is shown in \autoref{tab:ranker-hp}. Notably, we make the samples from $S_{\textrm{offline}}$ to be more likely to be sampled, because there are significantly more offline samples than online samples.

\begin{table}[t]
\centering
    \begin{tabular}{|l|c|}
    \hline
    Hidden units & 64 \\
    Learning rate ($\eta_{z}$ and $\eta_{\theta}$) & 0.0001 \\
    Optimizer & Adam \\
    Batch size & 4 \\
    $N_{\textrm{iters}}$ (batches per task) & 50 \\
    $N_{\textrm{trans}}$ (transf. iterations) & 20 \\
    $\lambda_{\textrm{sim}}$ (triplet loss weight) & 0.03 \\
    $\alpha$ (triplet loss margin) & 0.5 \\
    $\lambda_{\textrm{reg}}$ (SI weight) & 10000 \\
    $|S_{\textrm{offline}}| : |S_{\textrm{past}}| : |S|$ & 5 : 1 : 1 \\
    Exploration-exploitation ratio & 0.5 \\
    \hline
    \end{tabular}
\caption{Hyper-parameters to train the performance ranker. See Alg. 1 and \S 3 for explanation of notations.}
\label{tab:ranker-hp}
\end{table}

For configuration encoder used in NAS, we use GIN~\cite{xu2018powerful} with 2 layers, dropout rate 0.2 and a learn-able epsilon. We add two virtual nodes to aggregate information from all nodes in each layer, similar to \cite{siems2020bench}.

To accelerate the evaluation and save the huge computational cost to train and evaluate configurations repeatedly, we build a benchmark, \ie, a performance lookup table, which consists of the validation and test AP of a specific configuration trained on a specific dataset. For each dataset, we randomly sample 200 distinct configurations from HPO and NAS space respectively. Afterwards, the ranker is inferenced on the 200 and the best is selected from them. This practice follows many recent NAS works~\cite{wen2020neural,bender2018understanding,radosavovic2019network,klyuchnikov2020bench}. Notably, for larger space, the random sampling step can be easily replaced with an active learning approach (\eg, bayesian optimization), but since the newly sampled configuration is likely to be a unseen one, we would not be able to use a benchmark to accelerate this process.

\section{More experiment results}

The performance of \algname{} on each dataset is shown in \autoref{tab:performance-per-dataset}. Since the datasets can appear in random order during our evaluation, this table shows an average case of how much the dataset can benefit from others. \algname{} outcompetes baselines on most of the datasets. We also note that different datasets have different difficulties, causing performance gains to diverse. For example, we can easily get 1.1\% AP improvement on WIDER-360, but for WIKI, the baseline is very close to perfect and the room for improvement is very narrow. The standard deviations of multiple runs with different random seeds are also shown in those tables.

\begin{table*}[t]
\begin{subtable}[t]{\textwidth}
\centering
\resizebox{\textwidth}{!}{
\begin{tabular}{l|ccccccccccc|c}
\hline
               Method &                           AFLW &                          ANIME &                       FaceMask &                           FDDB &                       FDDB-360 &                           MAFA &                     Pascal VOC &                           UFDD &                       UMDAA-02 &                      WIDER-360 &                           WIKI &                        Average \\
\hline
      Random search &          \meanstd{99.15}{0.12} &          \meanstd{97.58}{0.18} &          \meanstd{94.32}{0.43} &          \meanstd{97.42}{0.22} &          \meanstd{97.00}{0.18} &          \meanstd{92.98}{0.48} &          \meanstd{97.36}{0.41} &          \meanstd{78.50}{0.59} &          \meanstd{99.63}{0.04} &          \meanstd{67.23}{0.97} &          \meanstd{99.67}{0.08} &          \meanstd{92.80}{9.90} \\
      Best on WIDER &          \meanstd{99.00}{0.00} &          \meanstd{97.53}{0.00} &          \meanstd{94.02}{0.00} &          \meanstd{97.11}{0.00} & \textbf{\meanstd{97.22}{0.00}} &          \meanstd{93.13}{0.00} &          \meanstd{97.12}{0.00} &          \meanstd{78.27}{0.00} &          \meanstd{99.60}{0.00} & \textbf{\meanstd{69.53}{0.00}} &          \meanstd{99.63}{0.00} &          \meanstd{92.92}{9.36} \\
          Tr-AutoML &          \meanstd{99.12}{0.12} &          \meanstd{97.58}{0.16} & \textbf{\meanstd{94.51}{0.35}} &          \meanstd{97.34}{0.29} &          \meanstd{96.91}{0.23} & \textbf{\meanstd{93.28}{0.35}} &          \meanstd{97.20}{0.60} &          \meanstd{78.59}{0.46} &          \meanstd{99.60}{0.06} &          \meanstd{66.84}{1.36} &          \meanstd{99.67}{0.06} &          \meanstd{92.79}{9.96} \\
          HyperSTAR &          \meanstd{99.09}{0.19} &          \meanstd{97.60}{0.16} &          \meanstd{94.32}{0.46} &          \meanstd{97.40}{0.25} &          \meanstd{96.97}{0.25} &          \meanstd{92.86}{0.40} &          \meanstd{97.21}{0.32} &          \meanstd{78.76}{0.42} &          \meanstd{99.63}{0.04} &          \meanstd{66.86}{1.23} &          \meanstd{99.66}{0.06} &          \meanstd{92.76}{9.94} \\
               SCoT &          \meanstd{99.17}{0.11} &          \meanstd{97.62}{0.20} &          \meanstd{94.25}{0.55} &          \meanstd{97.38}{0.16} &          \meanstd{96.91}{0.23} &          \meanstd{92.80}{0.46} &          \meanstd{97.39}{0.36} &          \meanstd{78.49}{0.57} &          \meanstd{99.62}{0.04} &          \meanstd{67.40}{0.72} &          \meanstd{99.69}{0.04} &          \meanstd{92.79}{9.86} \\
\hline
\algname{} (ResNet) &          \meanstd{99.21}{0.09} & \textbf{\meanstd{97.69}{0.14}} &          \meanstd{94.47}{0.32} &          \meanstd{97.53}{0.12} &          \meanstd{96.93}{0.21} &          \meanstd{92.68}{0.54} &          \meanstd{97.53}{0.39} &          \meanstd{78.63}{0.41} &          \meanstd{99.63}{0.03} &          \meanstd{66.59}{0.79} & \textbf{\meanstd{99.71}{0.04}} &         \meanstd{92.78}{10.05} \\
 \algname{} (stats) &          \meanstd{99.21}{0.06} &          \meanstd{97.61}{0.16} &          \meanstd{94.36}{0.36} &          \meanstd{97.54}{0.12} &          \meanstd{96.91}{0.15} &          \meanstd{92.83}{0.34} &          \meanstd{97.56}{0.12} & \textbf{\meanstd{78.77}{0.48}} &          \meanstd{99.62}{0.04} &          \meanstd{66.79}{1.07} &          \meanstd{99.70}{0.04} &          \meanstd{92.81}{9.98} \\
   \algname{} (MSE) &          \meanstd{99.17}{0.08} &          \meanstd{97.43}{0.20} &          \meanstd{94.43}{0.48} &          \meanstd{97.42}{0.24} &          \meanstd{96.96}{0.14} &          \meanstd{93.08}{0.46} &          \meanstd{97.34}{0.26} &          \meanstd{78.47}{0.58} &          \meanstd{99.62}{0.05} &          \meanstd{67.16}{0.80} &          \meanstd{99.69}{0.02} &          \meanstd{92.80}{9.91} \\
         \hline
\algname{} & \textbf{\meanstd{99.25}{0.01}} &          \meanstd{97.57}{0.11} &          \meanstd{94.39}{0.28} & \textbf{\meanstd{97.62}{0.02}} &          \meanstd{96.93}{0.22} &          \meanstd{92.82}{0.34} & \textbf{\meanstd{97.63}{0.05}} &          \meanstd{78.62}{0.32} & \textbf{\meanstd{99.64}{0.01}} &          \meanstd{68.31}{0.49} &          \meanstd{99.71}{0.01} & \textbf{\meanstd{92.95}{9.66}} \\
\hline
\end{tabular}
}
\caption{AP of searched configuration on HPO space. The higher the better.}
\label{tab:ap-per-dataset-hpo}
\end{subtable}

\hfill

\begin{subtable}[t]{\textwidth}
\centering
\resizebox{\textwidth}{!}{
\begin{tabular}{l|ccccccccccc|c}
\hline
               Method &                           AFLW &                          ANIME &                       FaceMask &                           FDDB &                       FDDB-360 &                           MAFA &                     Pascal VOC &                           UFDD &                       UMDAA-02 &                      WIDER-360 &                           WIKI &                         Average \\
\hline
      Random search &          \meanstd{99.07}{0.07} &          \meanstd{97.27}{0.19} &          \meanstd{93.95}{0.27} &          \meanstd{96.85}{0.26} &          \meanstd{96.70}{0.18} &          \meanstd{93.91}{0.29} &          \meanstd{95.58}{0.50} &          \meanstd{76.89}{0.60} &          \meanstd{99.72}{0.03} &          \meanstd{66.17}{0.43} &          \meanstd{99.51}{0.06} &          \meanstd{92.33}{10.24} \\
      Best on WIDER &          \meanstd{99.14}{0.00} &          \meanstd{97.37}{0.17} & \textbf{\meanstd{94.36}{0.09}} &          \meanstd{96.87}{0.08} & \textbf{\meanstd{96.95}{0.02}} &          \meanstd{93.65}{0.09} &          \meanstd{95.87}{0.00} &          \meanstd{77.39}{0.02} &          \meanstd{99.70}{0.01} & \textbf{\meanstd{66.75}{0.13}} & \textbf{\meanstd{99.59}{0.00}} &          \meanstd{92.51}{10.07} \\
          Tr-AutoML &          \meanstd{99.03}{0.13} &          \meanstd{97.38}{0.21} &          \meanstd{93.89}{0.42} &          \meanstd{96.71}{0.29} &          \meanstd{96.80}{0.11} &          \meanstd{93.95}{0.45} &          \meanstd{95.15}{0.76} &          \meanstd{76.67}{0.66} &          \meanstd{99.71}{0.04} &          \meanstd{66.35}{0.30} &          \meanstd{99.45}{0.11} &          \meanstd{92.28}{10.22} \\
          HyperSTAR &          \meanstd{99.06}{0.09} &          \meanstd{97.31}{0.18} &          \meanstd{93.99}{0.17} &          \meanstd{96.85}{0.25} &          \meanstd{96.71}{0.12} &          \meanstd{93.84}{0.34} &          \meanstd{95.65}{0.54} &          \meanstd{76.84}{0.63} &          \meanstd{99.71}{0.03} &          \meanstd{66.02}{0.39} &          \meanstd{99.50}{0.04} &          \meanstd{92.32}{10.29} \\
               SCoT &          \meanstd{99.07}{0.06} &          \meanstd{97.26}{0.16} &          \meanstd{93.93}{0.26} &          \meanstd{96.80}{0.27} &          \meanstd{96.71}{0.22} &          \meanstd{93.83}{0.33} &          \meanstd{95.53}{0.52} &          \meanstd{76.79}{0.62} &          \meanstd{99.73}{0.02} &          \meanstd{66.18}{0.35} &          \meanstd{99.50}{0.06} &          \meanstd{92.30}{10.25} \\
\hline
\algname{} (ResNet) &          \meanstd{99.13}{0.04} &          \meanstd{97.35}{0.13} &          \meanstd{94.07}{0.28} &          \meanstd{96.90}{0.14} &          \meanstd{96.78}{0.12} &          \meanstd{93.91}{0.33} &          \meanstd{95.79}{0.38} &          \meanstd{77.39}{0.24} &          \meanstd{99.73}{0.04} &          \meanstd{66.54}{0.21} &          \meanstd{99.57}{0.03} &          \meanstd{92.47}{10.11} \\
 \algname{} (stats) &          \meanstd{99.14}{0.05} &          \meanstd{97.33}{0.18} &          \meanstd{93.94}{0.15} &          \meanstd{96.93}{0.20} &          \meanstd{96.75}{0.15} &          \meanstd{94.03}{0.30} &          \meanstd{95.73}{0.47} &          \meanstd{77.41}{0.35} &          \meanstd{99.73}{0.03} &          \meanstd{66.63}{0.19} &          \meanstd{99.57}{0.03} &          \meanstd{92.47}{10.09} \\
   \algname{} (MSE) &          \meanstd{99.07}{0.07} &          \meanstd{97.21}{0.19} &          \meanstd{93.79}{0.19} & \textbf{\meanstd{97.10}{0.30}} &          \meanstd{96.67}{0.11} &          \meanstd{93.99}{0.22} &          \meanstd{95.38}{0.59} &          \meanstd{77.13}{0.36} & \textbf{\meanstd{99.73}{0.03}} &          \meanstd{66.18}{0.34} &          \meanstd{99.57}{0.03} &          \meanstd{92.35}{10.21} \\
         \hline
\algname{} & \textbf{\meanstd{99.16}{0.05}} & \textbf{\meanstd{97.40}{0.07}} &          \meanstd{93.89}{0.25} &          \meanstd{97.05}{0.15} &          \meanstd{96.90}{0.11} & \textbf{\meanstd{94.13}{0.05}} & \textbf{\meanstd{95.92}{0.12}} & \textbf{\meanstd{77.61}{0.07}} &          \meanstd{99.72}{0.03} &          \meanstd{66.64}{0.05} &          \meanstd{99.58}{0.02} & \textbf{\meanstd{92.55}{10.08}} \\
\hline
\end{tabular}
}
\caption{AP of searched configuration on NAS space. The higher the better.}
\label{tab:ap-per-dataset-nas}
\end{subtable}

\hfill

\begin{subtable}[t]{\textwidth}
\centering
\resizebox{\textwidth}{!}{
\begin{tabular}{l|ccccccccccc|c}
\hline
                   Method &                          AFLW &                          ANIME &                        FaceMask &                          FDDB &                      FDDB-360 &                           MAFA &                    Pascal VOC &                            UFDD &                       UMDAA-02 &                     WIDER-360 &                           WIKI &                        Average \\
\hline
          Random search &        \meanstd{20.09}{16.21} &         \meanstd{20.09}{16.22} &          \meanstd{20.09}{16.22} &        \meanstd{20.09}{16.22} &        \meanstd{20.09}{16.22} &         \meanstd{20.09}{16.21} &        \meanstd{20.09}{16.22} &          \meanstd{20.09}{16.22} &         \meanstd{20.09}{16.22} &        \meanstd{20.10}{16.21} &         \meanstd{20.09}{16.22} &         \meanstd{20.09}{16.21} \\
          Best on WIDER &         \meanstd{41.40}{0.00} &          \meanstd{19.91}{0.00} &           \meanstd{31.94}{0.00} &         \meanstd{47.22}{0.00} & \textbf{\meanstd{2.78}{0.00}} &          \meanstd{12.68}{0.00} &         \meanstd{30.09}{0.00} &           \meanstd{25.46}{0.00} &          \meanstd{30.09}{0.00} & \textbf{\meanstd{0.48}{0.00}} &          \meanstd{34.26}{0.00} &          \meanstd{25.12}{0.00} \\
              Tr-AutoML &        \meanstd{23.90}{17.57} &         \meanstd{19.87}{17.47} & \textbf{\meanstd{13.35}{11.52}} &        \meanstd{24.32}{19.19} &        \meanstd{27.39}{20.26} & \textbf{\meanstd{10.58}{9.94}} &        \meanstd{26.03}{22.64} &          \meanstd{17.81}{12.23} &         \meanstd{31.48}{22.02} &        \meanstd{29.49}{24.19} &         \meanstd{19.76}{15.84} &         \meanstd{22.18}{17.53} \\
              HyperSTAR &        \meanstd{26.49}{22.28} &         \meanstd{17.64}{14.24} &          \meanstd{20.14}{16.01} &        \meanstd{21.34}{18.66} &        \meanstd{21.60}{21.23} &         \meanstd{23.26}{14.44} &        \meanstd{27.18}{14.03} & \textbf{\meanstd{12.75}{10.95}} &         \meanstd{20.76}{15.37} &        \meanstd{27.76}{21.04} &         \meanstd{23.10}{15.07} &         \meanstd{22.00}{16.67} \\
                   SCoT &        \meanstd{16.56}{15.89} &         \meanstd{18.03}{15.29} &          \meanstd{23.77}{22.20} &        \meanstd{24.58}{14.01} &        \meanstd{26.90}{19.23} &         \meanstd{26.34}{15.43} &        \meanstd{19.10}{15.88} &          \meanstd{20.86}{16.35} &         \meanstd{21.94}{15.76} &        \meanstd{16.24}{11.10} &         \meanstd{14.84}{11.29} &         \meanstd{20.83}{15.68} \\
    \hline
\algname{} (ResNet) &        \meanstd{10.75}{12.58} & \textbf{\meanstd{10.09}{9.67}} &          \meanstd{14.23}{10.39} &         \meanstd{11.74}{9.95} &        \meanstd{26.41}{19.02} &         \meanstd{30.91}{19.29} &        \meanstd{12.72}{14.73} &          \meanstd{15.23}{12.01} &         \meanstd{20.22}{12.78} &        \meanstd{30.78}{15.75} & \textbf{\meanstd{8.41}{10.46}} &         \meanstd{17.41}{13.33} \\
\algname{} (Statistics) &         \meanstd{10.37}{9.27} &         \meanstd{16.00}{14.12} &          \meanstd{18.08}{13.88} &         \meanstd{10.42}{9.89} &        \meanstd{27.22}{15.80} &         \meanstd{24.53}{12.56} &         \meanstd{11.81}{5.32} &          \meanstd{13.26}{11.94} &         \meanstd{21.67}{16.47} &        \meanstd{28.55}{17.73} &         \meanstd{11.62}{10.50} &         \meanstd{17.59}{12.50} \\
       \algname{} (MSE) &        \meanstd{16.60}{11.28} &         \meanstd{35.40}{23.94} &          \meanstd{17.56}{15.89} &        \meanstd{20.19}{18.65} &        \meanstd{22.65}{14.64} &         \meanstd{17.15}{14.33} &        \meanstd{21.37}{11.39} &          \meanstd{18.48}{17.16} &         \meanstd{25.04}{18.04} &        \meanstd{20.18}{12.14} &          \meanstd{15.56}{7.39} &         \meanstd{20.93}{14.99} \\
             \hline
\algname{} & \textbf{\meanstd{4.48}{0.75}} &          \meanstd{18.20}{7.70} &           \meanstd{16.19}{8.62} & \textbf{\meanstd{3.83}{1.83}} &        \meanstd{25.75}{19.55} &         \meanstd{25.36}{12.19} & \textbf{\meanstd{8.32}{2.23}} &           \meanstd{15.35}{6.09} & \textbf{\meanstd{12.94}{5.15}} &          \meanstd{4.66}{4.09} &           \meanstd{9.63}{4.37} & \textbf{\meanstd{13.16}{6.60}} \\
\hline
\end{tabular}
}
\caption{Rank (normalized) of searched configuration on HPO space. The lower the better.}
\label{tab:rank-per-dataset-hpo}
\end{subtable}

\hfill

\begin{subtable}[t]{\textwidth}
\centering
\resizebox{\textwidth}{!}{
\begin{tabular}{l|ccccccccccc|c}
\hline
                   Method &                          AFLW &                         ANIME &                      FaceMask &                          FDDB &                      FDDB-360 &                          MAFA &                    Pascal VOC &                          UFDD &                        UMDAA-02 &                     WIDER-360 &                          WIKI &                       Average \\
\hline
          Random search &        \meanstd{20.10}{16.21} &        \meanstd{20.10}{16.21} &        \meanstd{20.10}{16.21} &        \meanstd{20.10}{16.21} &        \meanstd{20.10}{16.21} &        \meanstd{20.10}{16.21} &        \meanstd{20.10}{16.21} &        \meanstd{20.10}{16.21} &          \meanstd{20.10}{16.21} &        \meanstd{20.10}{16.20} &        \meanstd{20.10}{16.20} &        \meanstd{20.10}{16.21} \\
          Best on WIDER &          \meanstd{6.00}{0.00} &         \meanstd{11.04}{7.09} & \textbf{\meanstd{2.81}{1.27}} &         \meanstd{15.14}{3.53} & \textbf{\meanstd{2.62}{0.96}} &         \meanstd{34.91}{6.15} &          \meanstd{7.54}{0.00} &          \meanstd{4.51}{0.08} &           \meanstd{30.50}{4.69} & \textbf{\meanstd{2.64}{2.24}} & \textbf{\meanstd{3.01}{0.30}} &         \meanstd{10.97}{2.39} \\
              Tr-AutoML &        \meanstd{30.46}{29.48} &        \meanstd{16.94}{17.09} &        \meanstd{26.73}{26.91} &        \meanstd{27.82}{21.15} &          \meanstd{8.54}{7.09} &        \meanstd{21.39}{24.41} &        \meanstd{32.22}{27.95} &        \meanstd{30.82}{22.48} &          \meanstd{34.46}{21.25} &         \meanstd{13.47}{9.68} &        \meanstd{35.34}{29.03} &        \meanstd{25.29}{21.50} \\
              HyperSTAR &        \meanstd{23.70}{21.27} &        \meanstd{17.20}{14.12} &        \meanstd{15.05}{10.76} &        \meanstd{19.67}{15.01} &        \meanstd{18.15}{11.33} &        \meanstd{24.55}{18.37} &        \meanstd{18.02}{17.67} &        \meanstd{21.91}{17.90} &          \meanstd{26.15}{14.62} &        \meanstd{25.78}{14.72} &        \meanstd{21.83}{12.33} &        \meanstd{21.09}{15.28} \\
                   SCoT &        \meanstd{20.08}{13.80} &        \meanstd{20.12}{12.34} &        \meanstd{20.52}{16.06} &        \meanstd{23.53}{16.45} &        \meanstd{21.12}{18.48} &        \meanstd{25.43}{17.77} &        \meanstd{21.21}{18.01} &        \meanstd{22.69}{18.24} &           \meanstd{15.60}{8.85} &        \meanstd{19.02}{13.15} &        \meanstd{25.03}{19.44} &        \meanstd{21.30}{15.69} \\
    \hline
\algname{} (ResNet) &          \meanstd{9.03}{6.82} &         \meanstd{11.44}{6.42} &        \meanstd{14.34}{10.72} &         \meanstd{13.97}{6.63} &         \meanstd{11.75}{8.00} &        \meanstd{21.61}{16.19} &        \meanstd{13.10}{10.08} &          \meanstd{5.94}{5.18} &          \meanstd{17.31}{15.18} &          \meanstd{7.26}{6.16} &          \meanstd{5.90}{6.08} &         \meanstd{11.97}{8.86} \\
\algname{} (Statistics) &          \meanstd{8.03}{6.10} &        \meanstd{13.45}{15.37} &         \meanstd{17.42}{9.92} &        \meanstd{14.20}{10.38} &        \meanstd{14.82}{11.45} &        \meanstd{15.10}{14.21} &        \meanstd{15.85}{12.39} &          \meanstd{6.96}{4.78} & \textbf{\meanstd{15.43}{14.29}} &          \meanstd{4.82}{4.82} &          \meanstd{6.52}{6.74} &        \meanstd{12.06}{10.04} \\
       \algname{} (MSE) &        \meanstd{22.07}{19.74} &        \meanstd{26.07}{19.50} &        \meanstd{29.64}{15.64} &          \meanstd{8.93}{7.67} &        \meanstd{21.07}{11.73} &        \meanstd{15.08}{10.85} &        \meanstd{26.63}{21.48} &         \meanstd{12.42}{8.73} &          \meanstd{15.79}{12.35} &        \meanstd{18.69}{12.56} &          \meanstd{5.37}{6.05} &        \meanstd{18.34}{13.30} \\
             \hline
\algname{} & \textbf{\meanstd{5.04}{3.32}} & \textbf{\meanstd{6.73}{1.73}} &        \meanstd{23.15}{12.00} & \textbf{\meanstd{7.51}{6.50}} &          \meanstd{4.86}{4.19} & \textbf{\meanstd{5.42}{0.69}} & \textbf{\meanstd{6.83}{1.74}} & \textbf{\meanstd{3.03}{0.54}} &          \meanstd{16.70}{11.03} &          \meanstd{2.91}{0.49} &          \meanstd{3.46}{4.03} & \textbf{\meanstd{7.78}{4.21}} \\
\hline
\end{tabular}
}
\caption{Rank (normalized) of searched configuration on NAS space. The lower the better.}
\label{tab:rank-per-dataset-nas}
\end{subtable}

\caption{Performance of \algname{} per dataset, along with standard deviation.}
\label{tab:performance-per-dataset}

\end{table*}

\section{Analysis of performance benchmark}

As we have collected a large number of triplets (configurations, datasets and performances), we share some of the observations on this performance benchmark. We hope those insights will inspire future research work of transferable AutoML.

\begin{figure*}[t]
\centering
\includegraphics[width=\linewidth]{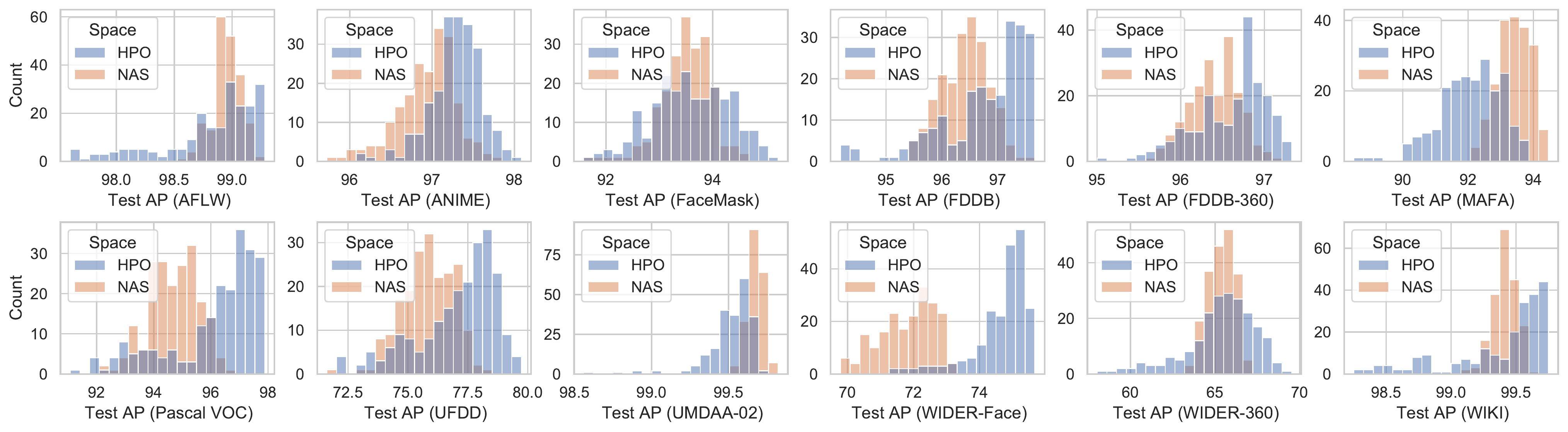}
\caption{AP distribution on 12 datasets. The bars count for how many configurations we have sampled lie in a specific AP range.}
\label{fig:ap-distribution}
\end{figure*}

\paragraph{Performance distribution and sensitivity to search space.}
We check the detection performance (AP@50), as shown in \autoref{fig:ap-distribution}. The hardest dataset is ``WIDER-360'', on which the best AP is less than 70\%. UMDAA-02 turns out to be the easiest dataset of all, on which the majority of configurations are above 99.2\%. The overall performance of HPO space is generally higher than NAS space, but MAFA and UMDAA-02 are two exceptions, where NAS is more useful than HPO.

The datasets that are most sensitive to hyper-parameter tuning are WIDER-360 and UFDD, where the gap between the best hyper-parameter and the worst differ by as much as 10\%. while performances on UMDAA-02, WIKI and AFLW are very close, with less than 2\% min-max-difference. For neural architecture search, the best and worst are closer in general, indicating that the final performance is less sensitive to changes in architectures alone.

To combine the advantages from both HPO and NAS space, one approach is do a joint search of hyper-parameters and neural architectures. However, this poses new challenges, both to search space design and search algorithms. There have been a few recent works that are jointly optimizing hyper-parameters and architectures~\cite{dong2020autohas,dai2020fbnetv3}, but this problem remains challenging and open, even outside the context of transferable AutoML.

\begin{figure}[t]
\centering
\includegraphics[width=\linewidth]{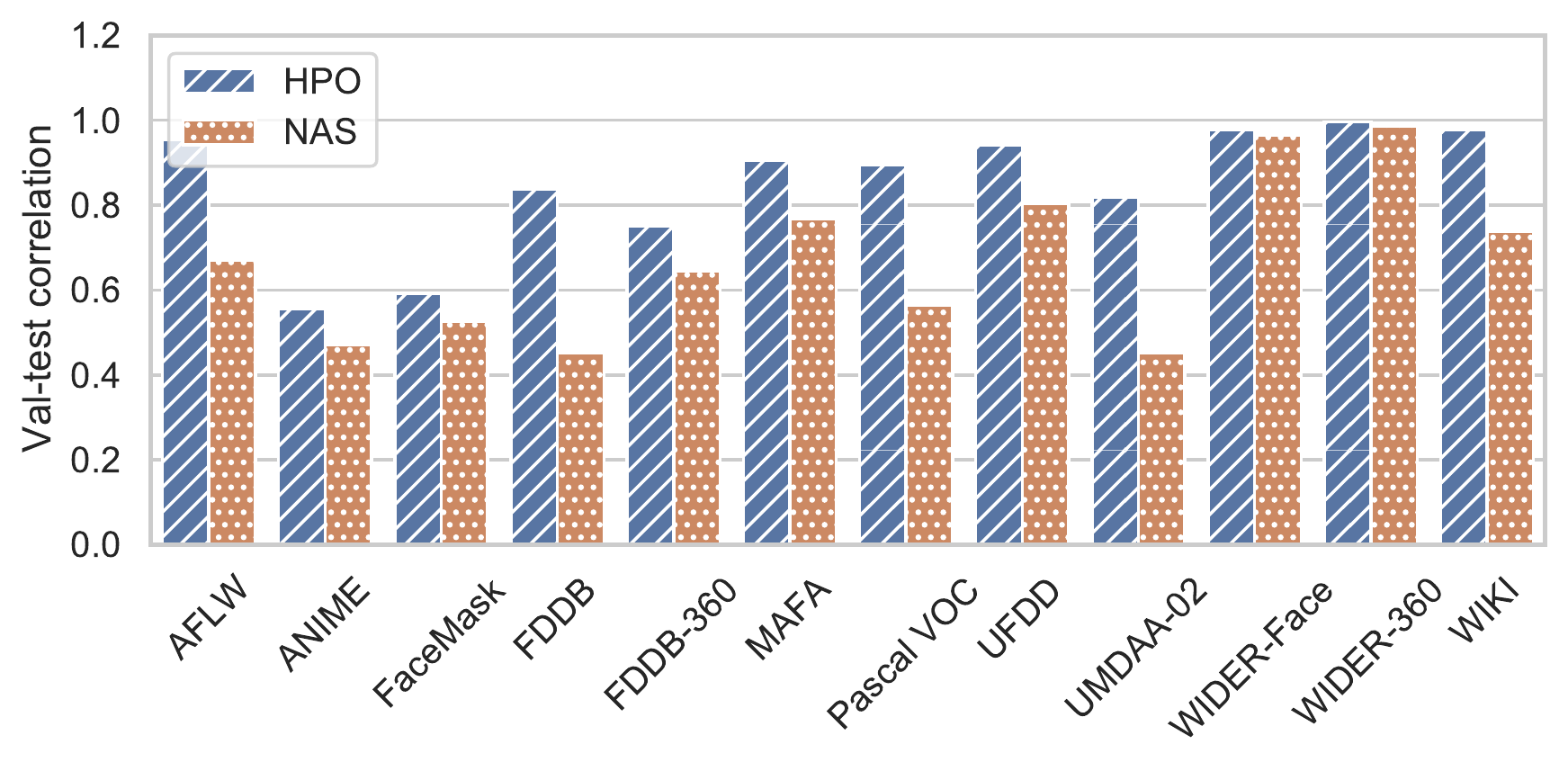}
\caption{Correlations of configuration rankings on validation dataset and test dataset.}
\label{fig:val-test-corr}
\end{figure}

\paragraph{Validation-test correlation.} For each dataset, we show the correlation between rankings on validation set versus rankings based on test set, \ie, whether the better configuration found with validation dataset still performs better on a unseen test dataset. The results are shown in \autoref{fig:val-test-corr}. If the correlation is low, it means that a model that performs better on validation set does not necessarily performs better on the test dataset. This could be caused by the gap between distributions of validation and test set, and indistinguishability between configurations. Datasets suffering from such problem is ANIME and FaceMask, and NAS space is generally worse than HPO space. However, most of the numbers (especially on HPO space) are still higher than 0.8, indicating that the trained model is still likely to perform well on a real-world unseen dataset.

\begin{figure}[t]
\centering
\begin{subfigure}[t]{\linewidth}
\centering
\includegraphics[width=.8\linewidth]{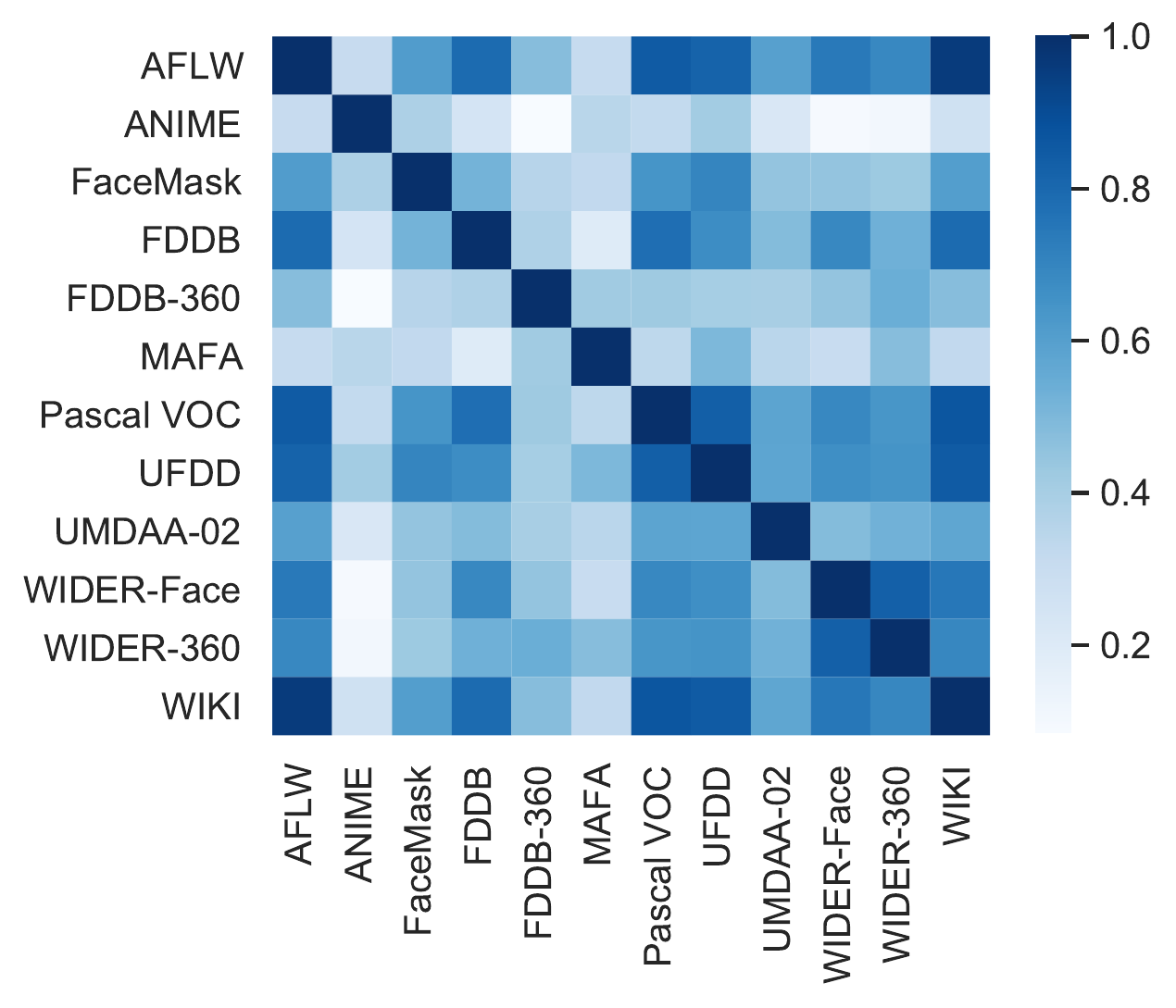}
\end{subfigure}
\begin{subfigure}[t]{\linewidth}
\centering
\includegraphics[width=.8\linewidth]{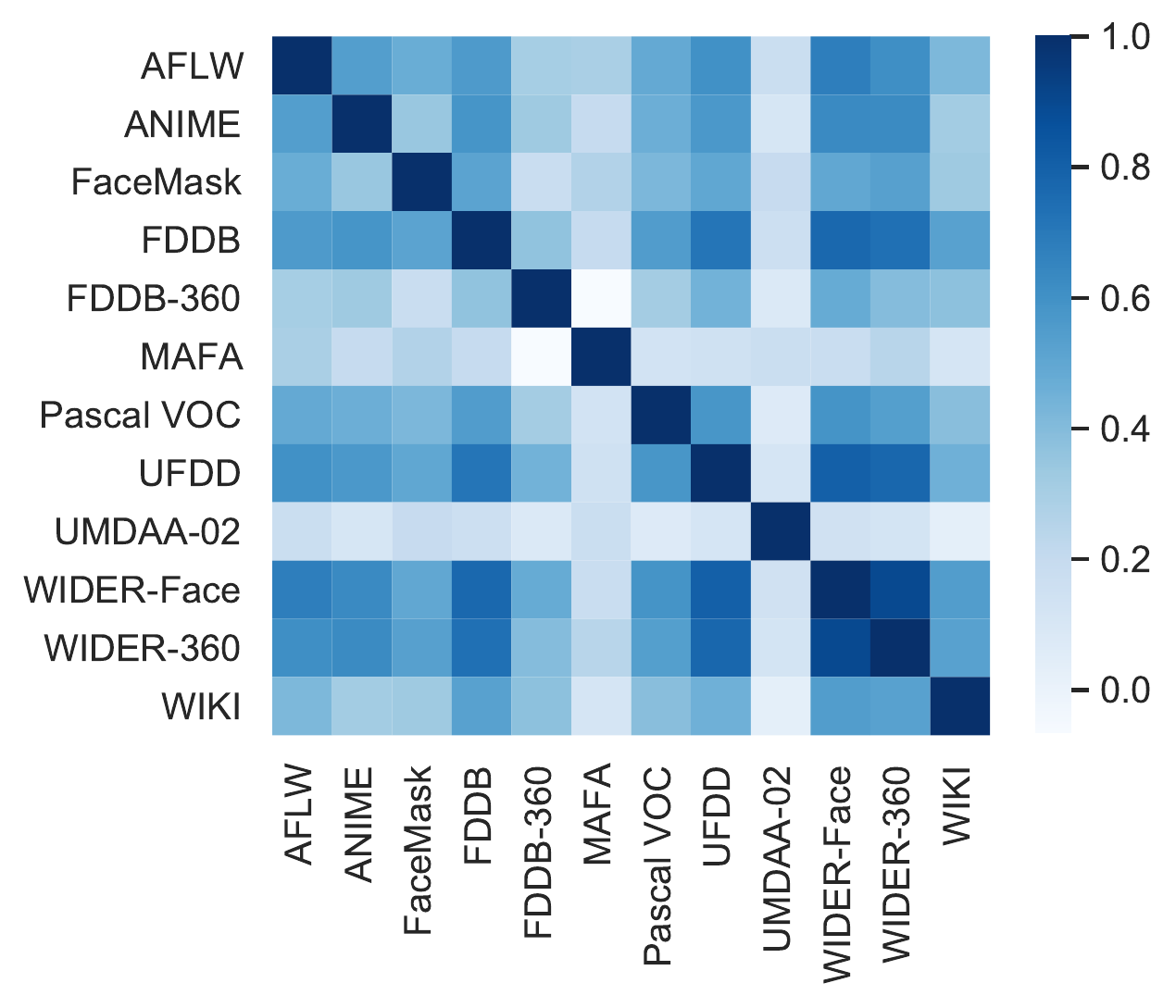}
\end{subfigure}
\caption{Mutual pearson correlation between hyperparameters on different datasets. The darker color indicates two datasets share more similar preferences for hyper-parameters / architectures. \textbf{(top)}: HPO space. \textbf{(bottom)}: NAS space.}
\label{fig:mutual-correlation}
\end{figure}

\paragraph{Configuration ranking correlation.} We examine the correlation between the ranking of configurations on different datasets, and check whether different datasets have similar favors to some configurations. The results are shown in \autoref{fig:mutual-correlation}. There are two interesting findings. (i) The heatmap on HPO and NAS space have a very different outlook, which means there is no transferability without a well-defined search space. For example, AFLW and WIKI have a high correlation on HPO space, but low on NAS. We hypothesis that the common characteristics that enables the transfer on HPO space does not apply well to architecture search. (ii) The overall correlations are higher for NAS space, indicating that datasets have more similar preferences for architectures. The correlations between WIDER-Face and WIDER-360 are especially high, which explains why ``Best on WIDER'' outperforms all other methods in \autoref{tab:performance-per-dataset}.

\begin{figure*}[htbp]
    \centering
    \begin{subfigure}{0.48\linewidth}
        \centering
        \includegraphics[width=0.95\linewidth]{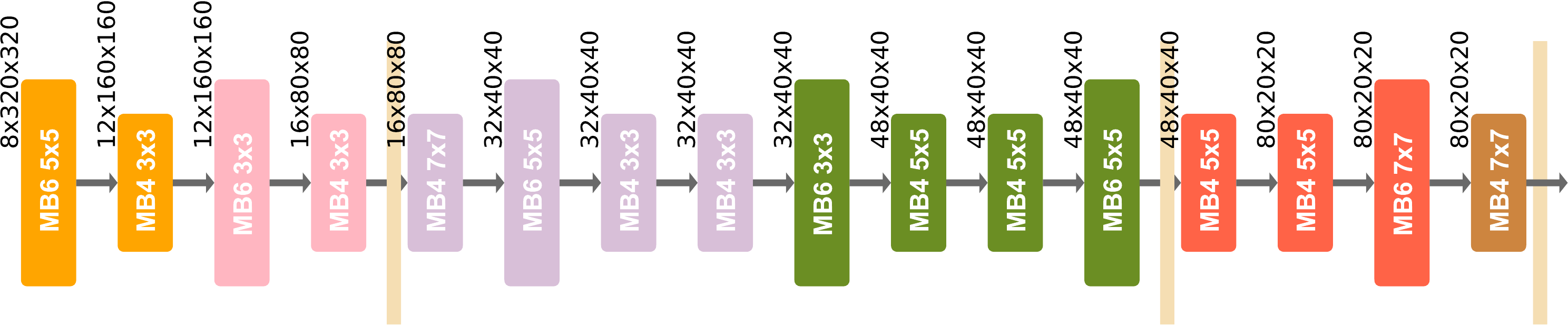}
        \caption{AFLW}
    \end{subfigure}
    \begin{subfigure}{0.48\linewidth}
        \centering
        \includegraphics[width=0.95\linewidth]{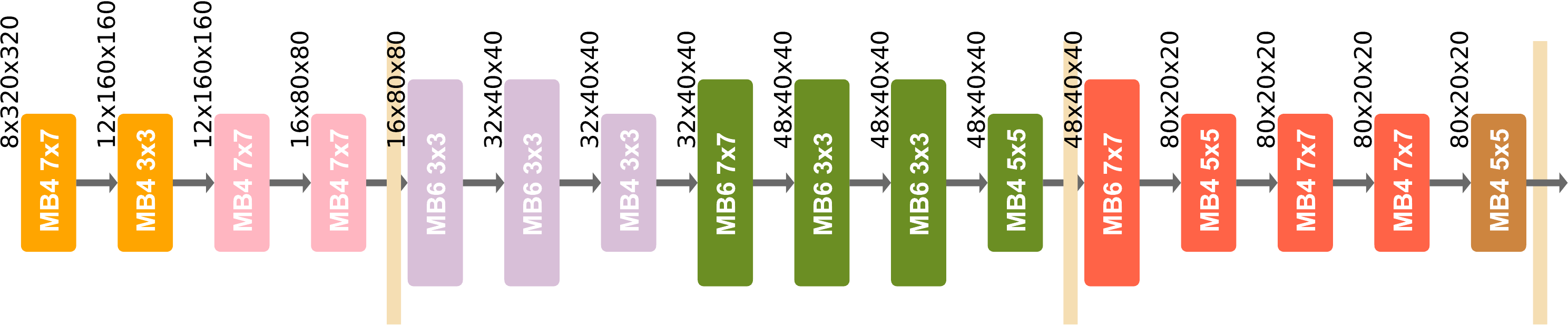}
        \caption{ANIME}
    \end{subfigure}
    \begin{subfigure}{0.48\linewidth}
        \centering
        \includegraphics[width=0.95\linewidth]{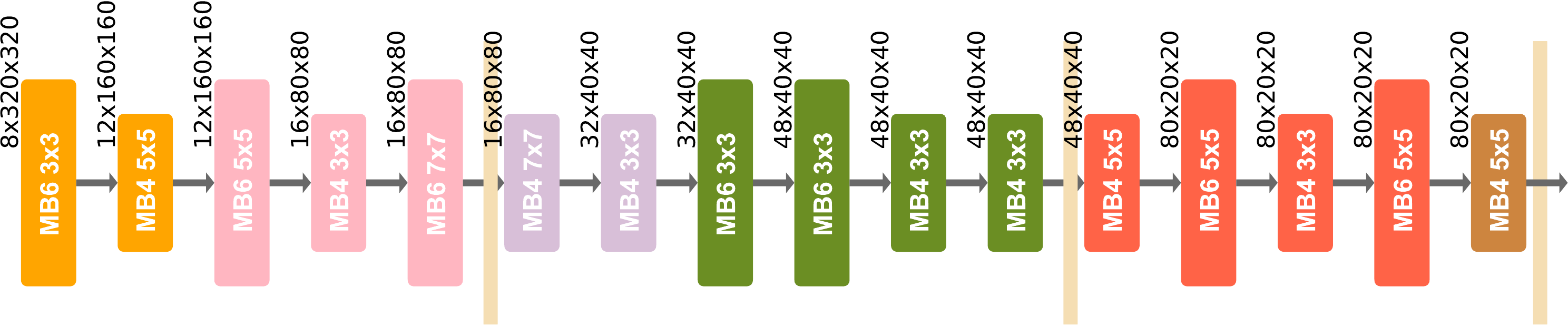}
        \caption{FaceMask}
    \end{subfigure}
    \begin{subfigure}{0.48\linewidth}
        \centering
        \includegraphics[width=0.95\linewidth]{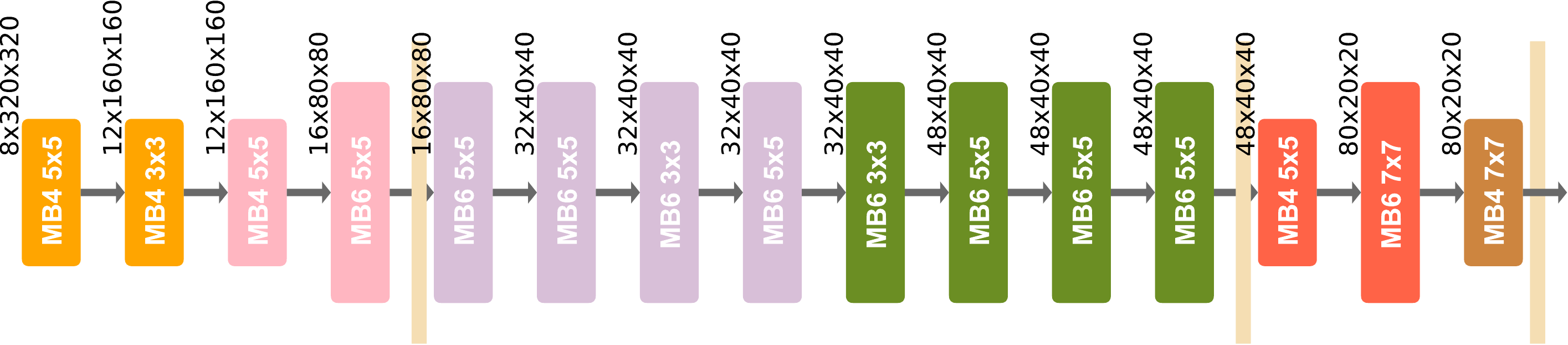}
        \caption{FDDB-360}
    \end{subfigure}
    \begin{subfigure}{0.48\linewidth}
        \centering
        \includegraphics[width=0.95\linewidth]{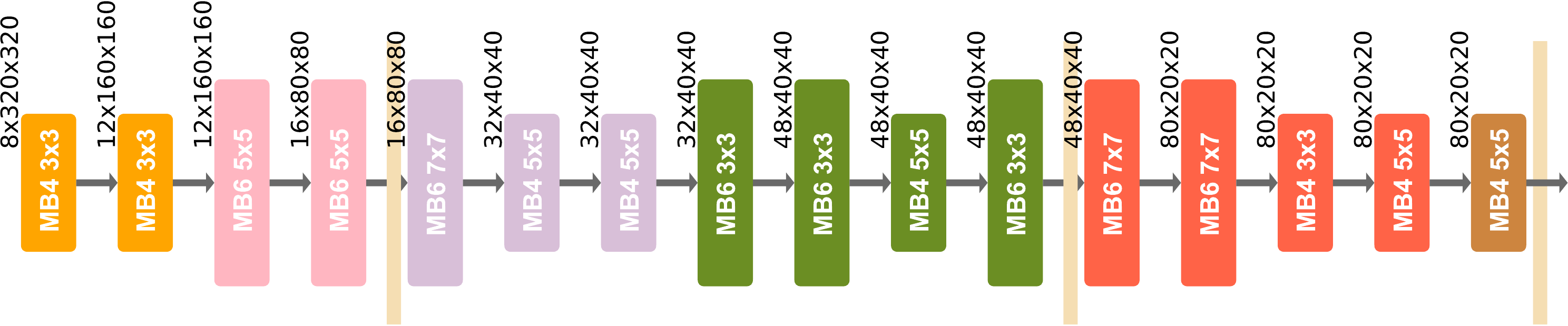}
        \caption{FDDB}
    \end{subfigure}
    \begin{subfigure}{0.48\linewidth}
        \centering
        \includegraphics[width=0.95\linewidth]{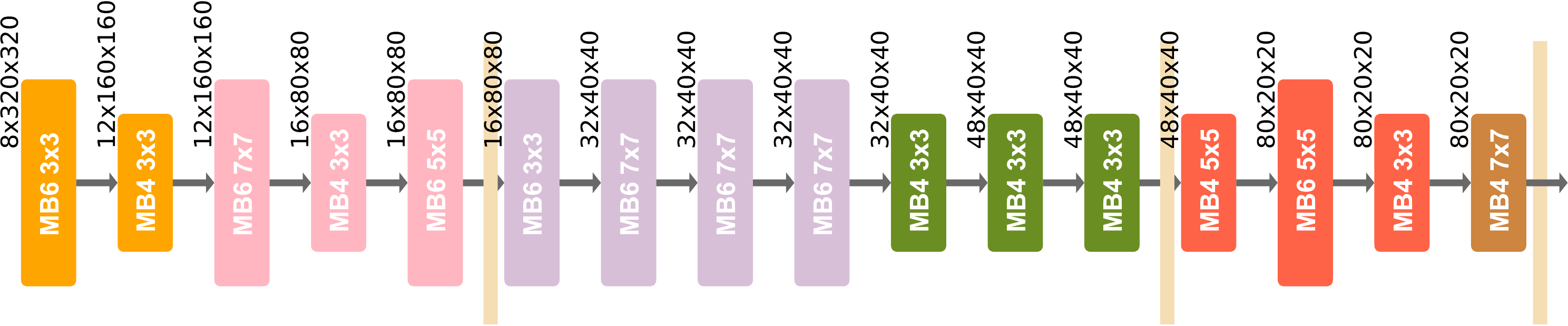}
        \caption{MAFA}
    \end{subfigure}
    \begin{subfigure}{0.48\linewidth}
        \centering
        \includegraphics[width=0.95\linewidth]{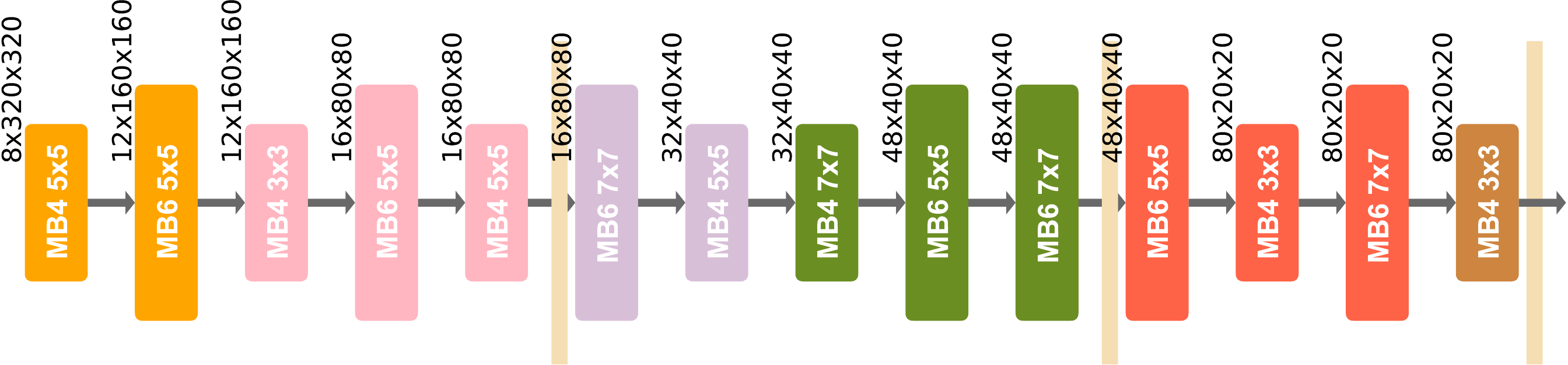}
        \caption{Pascal VOC}
    \end{subfigure}
    \begin{subfigure}{0.48\linewidth}
        \centering
        \includegraphics[width=0.95\linewidth]{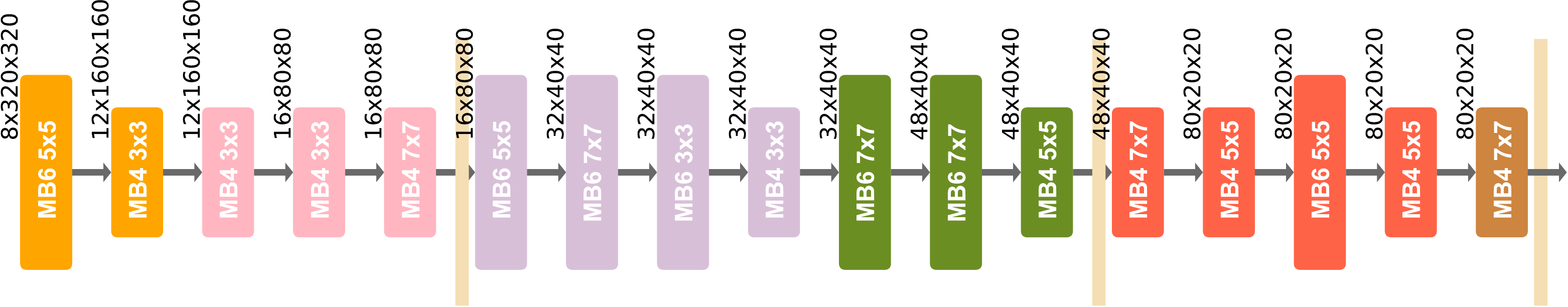}
        \caption{UFDD}
    \end{subfigure}
    \begin{subfigure}{0.48\linewidth}
        \centering
        \includegraphics[width=0.95\linewidth]{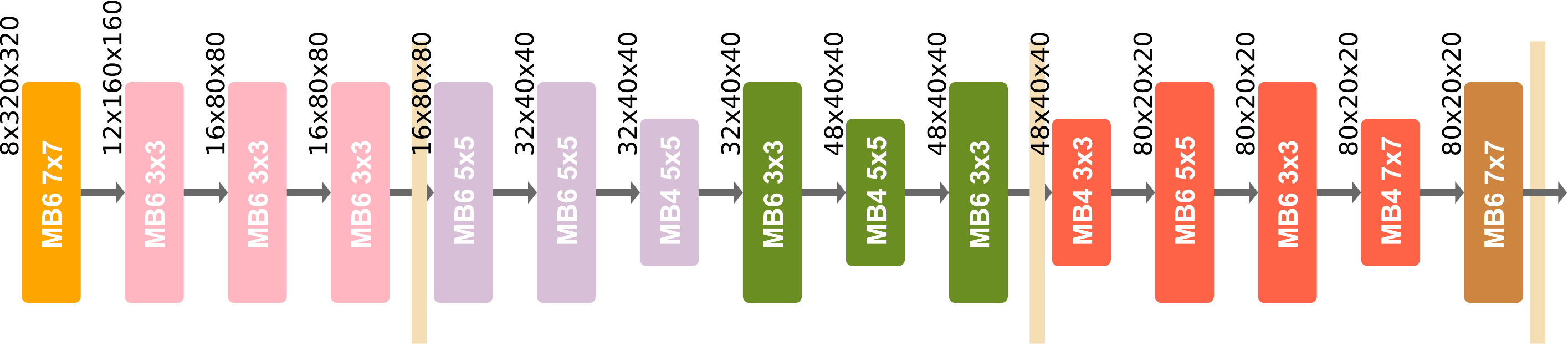}
        \caption{UMDAA-02}
    \end{subfigure}
    \begin{subfigure}{0.48\linewidth}
        \centering
        \includegraphics[width=0.95\linewidth]{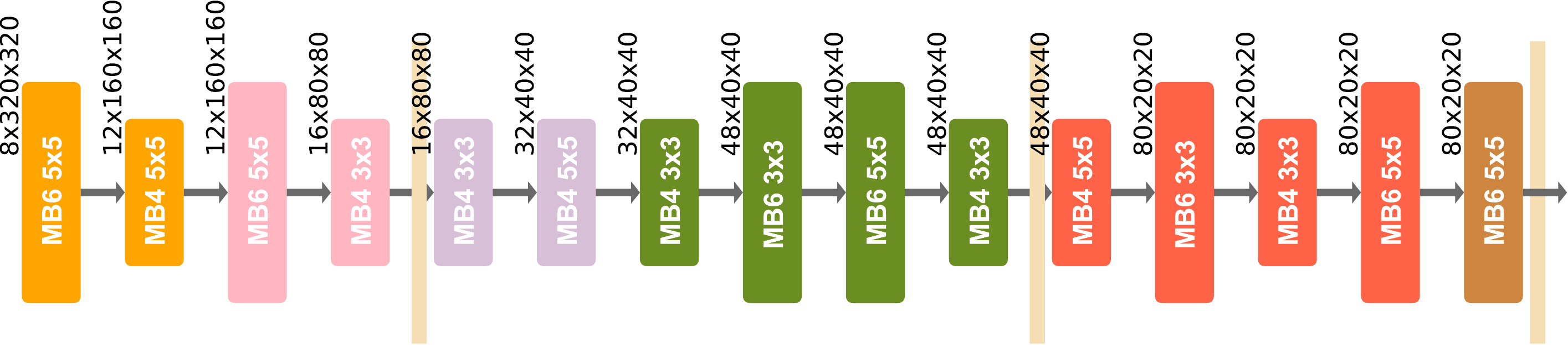}
        \caption{WIDER-Face}
    \end{subfigure}
    \begin{subfigure}{0.48\linewidth}
        \centering
        \includegraphics[width=0.95\linewidth]{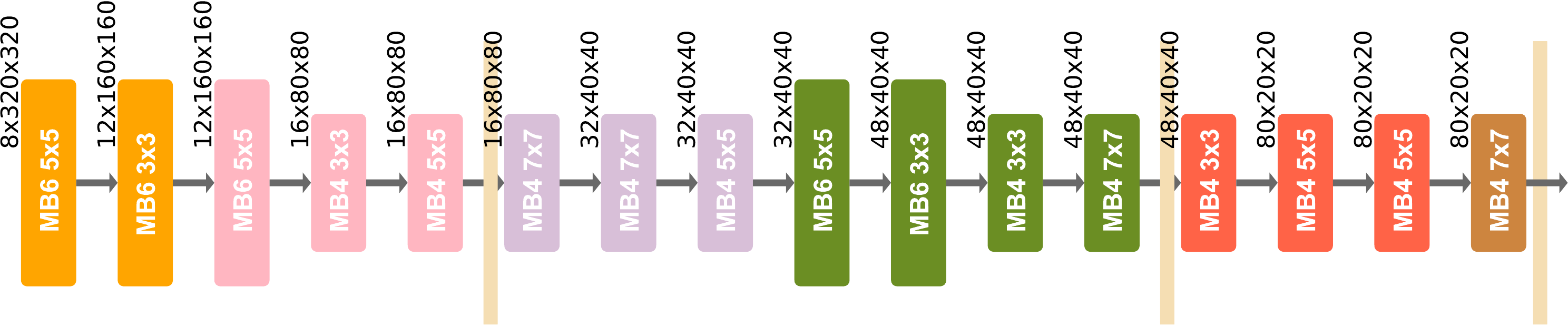}
        \caption{WIDER-360}
    \end{subfigure}
    \begin{subfigure}{0.48\linewidth}
        \centering
        \includegraphics[width=0.95\linewidth]{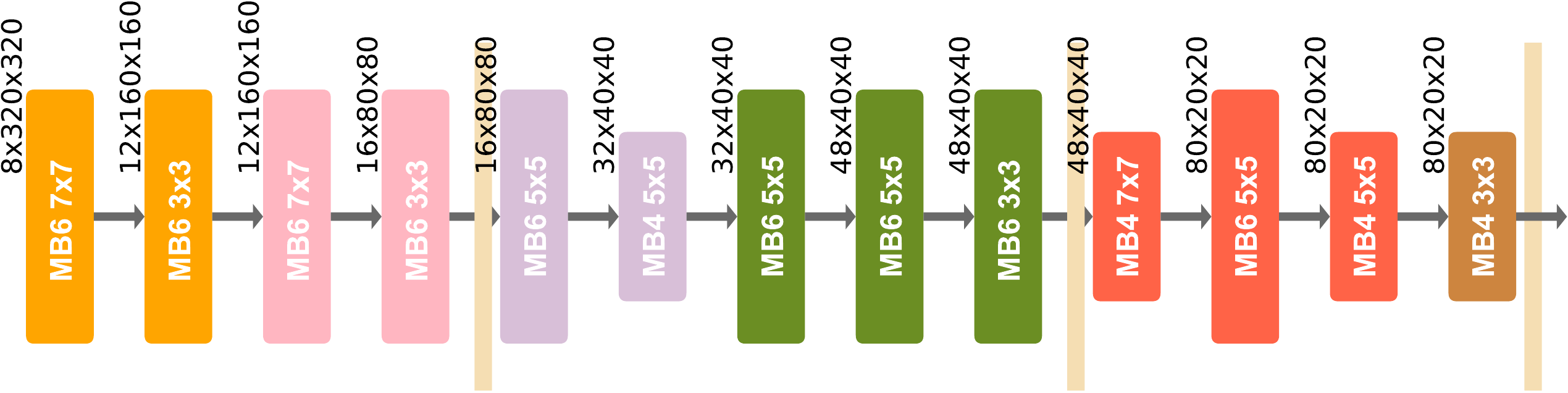}
        \caption{WIKI}
    \end{subfigure}
    \caption{Best neural architecture (backbone) on each dataset.}
    \label{fig:best-architecture}
\end{figure*}

\paragraph{Best hyper-parameter/architecture visualization.} We show the best hyper-parameter found on our search space in \autoref{tab:best-hyper-parameter}, and the best architecture found in \autoref{fig:best-architecture}. As illustrated, we observe no clear similarities among the best hyper-parameters or architectures, suggesting the need of tailored hparam/arch for each dataset.

\begin{table}[htbp]
\begin{adjustbox}{width=\linewidth}
\begin{tabular}{l|l}
\hline
Dataset    & Best hyper-parameter                                             \\
\hline
AFLW       & crop:0.55\_iou:0.5\_locw:8.0\_negp:2.0\_lr:3e-04\_sgd  \\
ANIME      & crop:0.55\_iou:0.4\_locw:2.0\_negp:2.0\_lr:1e-03\_adam \\
FaceMask   & crop:0.3\_iou:0.4\_locw:2.0\_negp:7.0\_lr:3e-04\_adam  \\
FDDB       & crop:0.55\_iou:0.5\_locw:8.0\_negp:2.0\_lr:3e-04\_sgd  \\
FDDB-360   & crop:0.55\_iou:0.5\_locw:8.0\_negp:7.0\_lr:1e-03\_adam \\
MAFA       & crop:0.55\_iou:0.5\_locw:2.0\_negp:2.0\_lr:3e-04\_adam \\
Pascal VOC & crop:0.3\_iou:0.4\_locw:8.0\_negp:2.0\_lr:3e-04\_adam  \\
UFDD       & crop:0.3\_iou:0.4\_locw:2.0\_negp:2.0\_lr:3e-03\_sgd   \\
UMDAA-02   & crop:0.55\_iou:0.5\_locw:4.0\_negp:7.0\_lr:1e-03\_sgd  \\
WIDER-Face      & crop:0.3\_iou:0.6\_locw:2.0\_negp:7.0\_lr:3e-03\_sgd   \\
WIDER-360  & crop:0.55\_iou:0.6\_locw:2.0\_negp:7.0\_lr:3e-03\_sgd` \\
WIKI       & crop:0.55\_iou:0.4\_locw:4.0\_negp:2.0\_lr:1e-03\_sgd \\
\hline
\end{tabular}
\end{adjustbox}
\caption{Best hyper-parameter searched on each dataset.}
\label{tab:best-hyper-parameter}
\end{table}

\end{document}